\documentclass[10pt,twocolumn,letterpaper]{article}

\usepackage[pagenumbers]{wacv} 

\usepackage{epsfig}
\usepackage{lipsum} 
\usepackage{multirow}
\usepackage{colortbl}
\usepackage{arydshln}
\usepackage{adjustbox}
\usepackage{caption}
\usepackage{graphicx}
\usepackage{amsmath}
\usepackage{amssymb}
\usepackage{booktabs}
\usepackage{float}
\usepackage[dvipsnames]{xcolor}
\usepackage{xcolor, soul}
\sethlcolor{yellow}
\usepackage{balance}
\usepackage[accsupp]{axessibility}  

\usepackage{xcolor}
\colorlet{darkgreen}{green!65!black}
\colorlet{darkblue}{blue!75!black}
\colorlet{darkred}{red!80!black}
\definecolor{lightblue}{HTML}{0071bc}
\definecolor{lightgreen}{HTML}{39b54a}
\definecolor{manyshot}{HTML}{6969ff}
\definecolor{medshot}{HTML}{f7c600}
\definecolor{fewshot}{HTML}{ff6969}
\definecolor{mypurple}{HTML}{412F8A}
\definecolor{myorange}{HTML}{fc8e62}
\definecolor{carolinablue}{rgb}{0.6, 0.73, 0.89}
\definecolor{indigo(web)}{rgb}{0.29, 0.0, 0.51}

\definecolor{deemph}{gray}{0.55}

\usepackage[pagebackref,breaklinks,colorlinks]{hyperref}

\renewcommand{\paragraph}[1]{\vspace{1.25mm}\noindent\textbf{#1}}
\newcommand{\grayrow}{\rowcolor[gray]{.9}}
\definecolor{baselinecolor}{gray}{.95}

\makeatletter
\def\blfootnote{\xdef\@thefnmark{}\@footnotetext}
\makeatother

\usepackage[capitalize]{cleveref}
\crefname{section}{Sec.}{Secs.}
\Crefname{section}{Section}{Sections}
\Crefname{table}{Table}{Tables}
\crefname{table}{Tab.}{Tabs.}

\begin{document}

\title{Motion Matters: Neural Motion Transfer for \\ Better Camera Physiological Measurement}

\author{Akshay Paruchuri$^{1}$, Xin Liu$^{2}$, Yulu Pan$^{1}$, Shwetak Patel$^{2}$, Daniel McDuff$^{2,*}$, Soumyadip Sengupta$^{1,*}$\\
$^{1}$UNC Chapel Hill\hspace{1em}$^{2}$University of Washington\\
{\tt\small \{akshay, ronisen, yulupan\}@cs.unc.edu, \{xliu0, shwetak, dmcduff\}@cs.washington.edu}
}
\maketitle

\blfootnote{$^{*}$denotes equal advising}   

\begin{abstract}
    Machine learning models for camera-based physiological measurement can have weak generalization due to a lack of representative training data. Body motion is one of the most significant sources of noise when attempting to recover the subtle cardiac pulse from a video. We explore motion transfer as a form of data augmentation to introduce motion variation while preserving physiological changes of interest. We adapt a neural video synthesis approach to augment videos for the task of remote photoplethysmography (rPPG) and study the effects of motion augmentation with respect to 1) the magnitude and 2) the type of motion. After training on motion-augmented versions of publicly available datasets, we demonstrate a 47\% improvement over existing inter-dataset results using various state-of-the-art methods on the PURE dataset. We also present inter-dataset results on five benchmark datasets to show improvements of up to 79\% using TS-CAN, a neural rPPG estimation method. Our findings illustrate the usefulness of motion transfer as a data augmentation technique for improving the generalization of models for camera-based physiological sensing. We release our code for using motion transfer as a data augmentation technique on three publicly available datasets, UBFC-rPPG, PURE, and SCAMPS, and models pre-trained on motion-augmented data here: \url{https://motion-matters.github.io/}
\end{abstract}

\vspace{-1em}
\section{Introduction}
\vspace{-0.5em}
\label{sec:intro}

Scalable health sensors enable frequent, opportunistic, and more equitable access to vital information about the body's internal state. Cameras are some of the most versatile and widely available sensors. Videos capture spatial, temporal, and ultimately frequency-specific information making them suitable for imaging dynamic processes, even below the surface of the skin~\cite{nowara2022seeing}. Camera-based measurement of cardiac signals is one such application~\cite{mcduff2021camera}, in which cameras are used to measure the pulse via light reflected from the body, a principle known as photoplethysmography (PPG)~\cite{blazek2000near,verkruysse2008remote}. The PPG signals can be used to derive respiration~\cite{poh2010advancements}, heart rate variability~\cite{poh2010advancements}, arrhythmia~\cite{poh2018diagnostic}, and blood pressure~\cite{jeong2016introducing}. As a result this technology has the potential to turn webcams and smartphones into meaningful health sensors.

\begin{figure}[t]
    \begin{center}
    \includegraphics[width=3in]{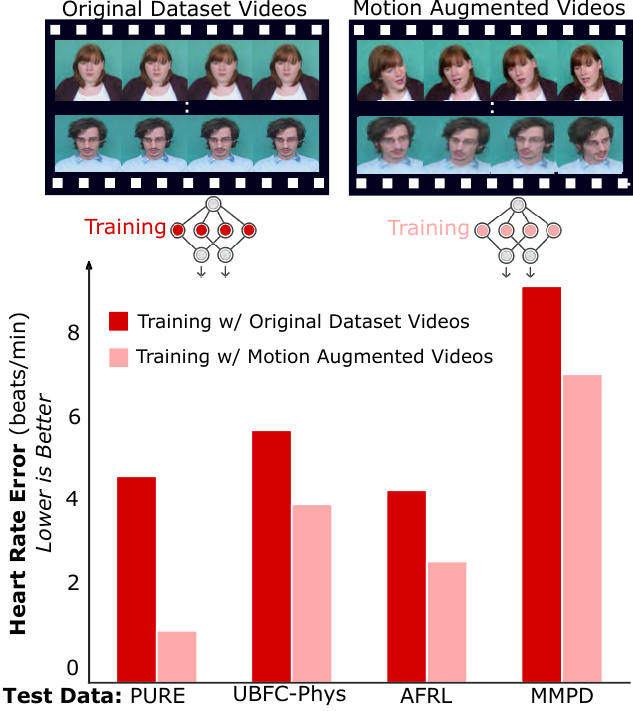}
    \end{center}
    \vspace{-0.5em}
    \caption{\textbf{Motion augmentation improves rPPG.} We present the first neural motion augmentation pipeline for the task of remote PPG estimation and empirically show it reduces error in heart rate estimation by up to 79\% in inter-dataset results using TS-CAN and 47\% over existing results using SOTA methods on PURE.}
    \label{fig:teaser}
    \vspace{-2em}
    \end{figure}

However, unlike traditional medical sensors, extracting physiological signals from a video requires more than filtering and simple signal processing. The state-of-the-art (SOTA) algorithms are supervised neural models~\cite{chen2018deepphys,vspetlik2018visual,yu2019remote,liu2020multi,yu2021physformer}. Despite the prowess of these models, they are inherently limited by the diversity of the data used to train them.
Public datasets (e.g., UBFC-rPPG~\cite{bobbia2019unsupervised}, PURE~\cite{stricker2014non}) serve as an extremely valuable resource for the research community, containing videos and synchronized physiological gold-standard measurements making them suitable for training and testing models. Building datasets such as these is challenging for two reasons: (1) collecting videos with gold-standard signals from a medical-grade sensor is time consuming and labor intensive, (2) it requires storing and distributing privacy sensitive biometric data. Therefore, more data efficient methods for training rPPG sensing models would be desirable. 

Synthetic data are a powerful resource in machine learning. The two main sources of synthetic data are (1) parametric computer graphics engines and (2) statistically-based generative machine learning models. Data created using these approaches have been used successfully for many computer vision tasks, including face detection, landmark localization, face
parsing and face recognition~\cite{mcduff2018identifying,wood2021fake,mcduff2019characterizing}, body pose estimation~\cite{shotton2011real} and eye tracking~\cite{wood2015rendering,swirski2014rendering}.

However, creating synthetic data that preserve the subtle and nuanced peripheral pulse in a video is non-trivial. McDuff et al.~\cite{mcduff2022scamps} released a large dataset (2,800 videos) of avatars and cardiac signals; however, their computer graphics pipeline had an extremely high computational overhead. 
Wang et al.~\cite{wang2022synthetic} used a learning based method to generate synthetic videos given a reference image and target PPG signal. Their creative approach successfully incorporated PPG signals to produce videos that benefited training. However, the videos created lacked the visual fidelity of other synthetics or real video datasets, and their pipeline involved several relatively complex components.

\begin{figure*}[ht]
    \begin{center}
    \includegraphics[width=6.5in]{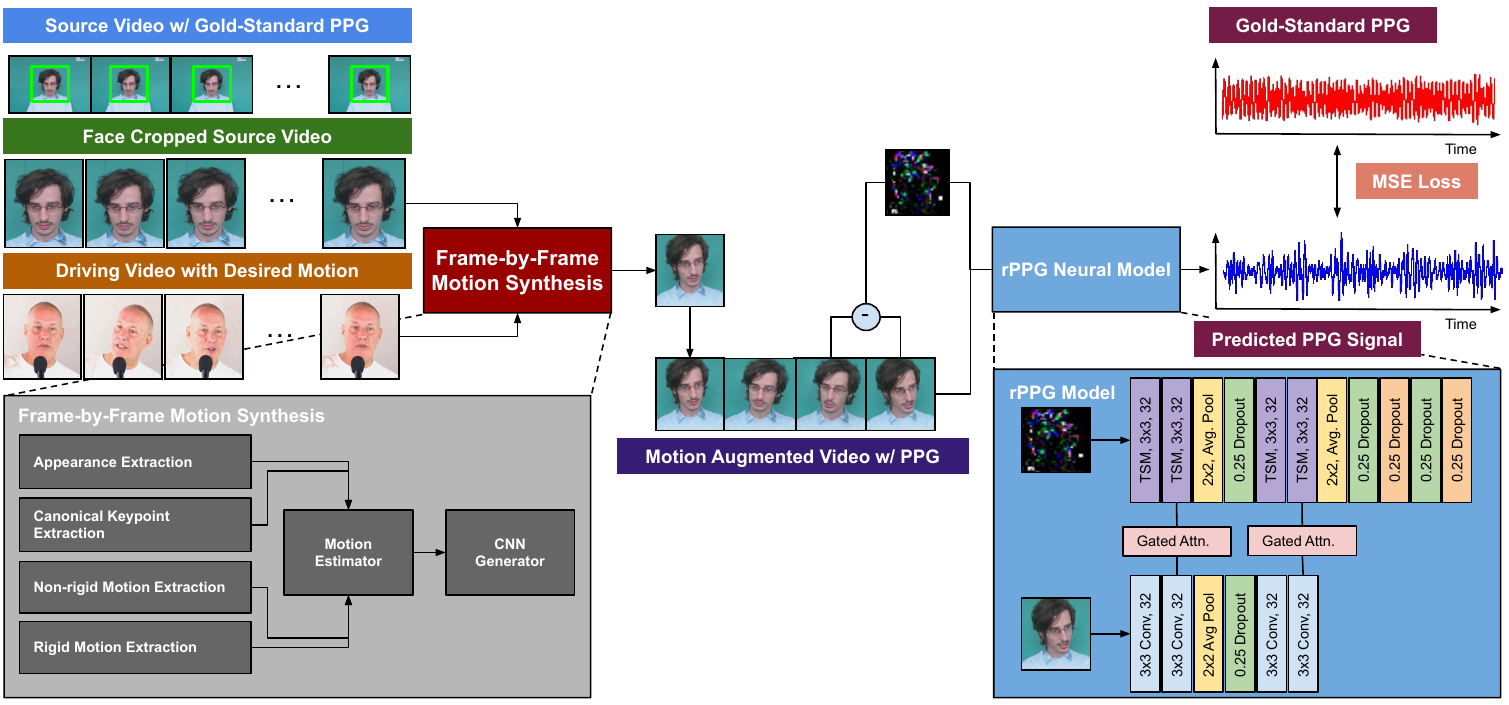}
    \end{center}
    \vspace{-0.5cm}
    \caption{\small{\textbf{Motion augmentation and training pipeline.} We augment frames of a source video with corresponding frames of a selected driving video to create an augmented video with the identity of the source video and motion of the target driving video. We then train a remote PPG estimation network on the augmented video with a mean squared error (MSE) loss.}}
    \label{fig:pipeline}
    \vspace{-0.35cm}
    \end{figure*}

We question whether existing motion transfer algorithms can be used effectively for augmenting rPPG video data and explore what steps need to be taken to achieve optimal results.
Our main contributions are as follows: 
\vspace{-0.1cm}
\begin{itemize}
  \item A systematic investigation of the impact of motion augmentation on the physiological information within rPPG videos.
  \vspace{-0.2cm}
  \item Quantitative, empirical evidence that conveys the meaningfulness of training with motion-augmented data, including (1) the benefits of different kinds of motion-augmented data, (2) consistent motion augmentations across neural motion transfer algorithms, (3) the benefit of naturalistic head motion over other synthetic methods (e.g., SCAMPS), and (4) consistent improvements using motion-augmented data regardless of a chosen rPPG estimation model.
  \vspace{-0.2cm}
  \item Our motion augmentation pipeline, using which in Table~\ref{tab:sota_pure} we demonstrate that our approach surpasses the SOTA when compared to other methods that train on UBFC-rPPG~\cite{bobbia2019unsupervised} and test on PURE~\cite{stricker2014non}. We also provide comprehensive inter-dataset results (see \Cref{tab:main_overall_results}) that highlight the usefulness of motion augmentation for improving the generalization of models for camera-based physiological sensing.
\end{itemize}

We summarize the key findings of this paper about the effectiveness of motion transfer as a data augmentation tool in \Cref{sec:disc}. We provide our code for augmenting datasets, training using these data, and pre-trained models trained on motion-augmented data (all assets are released with responsible use licenses~\cite{contractor2022behavioral}).

\vspace{-0.5em}
\section{Background}
\vspace{-0.5em}

\textbf{Generative Synthetics for Training Models:}
Statistical generative models~\cite{goodfellow2020generative,kingma2018glow,sohl2015deep,ho2020denoising,dhariwal2021diffusion} capture a probabilistic representation of a dataset from which samples can be drawn. These models are typically trained to mimic the distribution of the training set and can be trained without the need for labels, allowing large sets of data to be used.
Facial video generation using generative models has advanced rapidly over recent years~\cite{Liu2020GenerativeAN,Sha2021DeepPG}. Numerous image-driven works have accomplished the ability to separate identity and pose in source and driving images used for high quality, robust video generation using generative adversarial networks (GANs)~\cite{zakharov2019few, Siarohin_2019_NeurIPS,Wang2020OneShotFN,Hong2022DepthAwareGA}. Image-driven facial video generation methods attempt to preserve the identity of a given source image while manipulating the pose based on a driving video to generate a new video. The identity from the driving video is excluded with the help of a keypoint-based motion transfer approach, where keypoints are predicted for both a source image and a driving image in order to model local motion using shifts in the corresponding keypoints~\cite{Siarohin_2019_NeurIPS,Wang2020OneShotFN,Hong2022DepthAwareGA}. Face video generation that is achieved by using keypoints that take pose and expression into account can be successful for the task of head video generation, but can at times have a loss in source image identity and unwanted temporal artifacts~\cite{Siarohin_2019_NeurIPS,zakharov2019few,Hong2022DepthAwareGA}. Face-Vid2Vid~\cite{Wang2020OneShotFN} utilizes canonical keypoints in addition to source and driving image keypoints in order to capture a target person’s geometry signature, which includes the shape of the target's face, nose, and eyes. This allows for improved head video generation that minimizes source identity loss while effectively transferring motion from a driving video.

\textbf{rPPG Models:} The principle that photoplethysmography could be performed with a camera and without contact with the body was established by Blazek et al.~\cite{blazek2000near} and replicated in a series of following experiments~\cite{takano2007heart,verkruysse2008remote}. The application of more advanced signal processing methods helped make measurement somewhat more robust under real-world conditions~\cite{poh2010advancements,wang2017algorithmic}, as did leveraging knowledge of physiological and physical properties~\cite{wang2017algorithmic}. Yet, these models were still very sensitive to body motions. Both task-specific and multi-task neural, data-driven models currently achieve SOTA results in most cases~\cite{chen2018deepphys,yu2019remote,liu2020multi,yu2021physformer,liu2021efficientphys, narayanswamy2023bigsmall}, but are a function of the data used to train them. While intra-dataset performance is generally strong, inter-dataset performance is often substantively worse. In order to alleviate the dependency on labeled data, several researchers have proposed unsupervised learning procedures~\cite{gideon2021way,wang2022self,yang2022simper,Speth_2023_CVPR}. However, most require fine-tuning on a labeled set and also reveal that supervised learning still holds some additional benefit.
As an alternative or a complement, generative methods have been suggested to ``create'' data~\cite{mcduff2021synthetic,wang2022synthetic}.

\textbf{rPPG Datasets:} As with many health applications, those working in camera physiological measurement face challenges associated with collecting and managing data.  Public datasets (such as UBFC-rPPG~\cite{bobbia2019unsupervised}, PURE~\cite{stricker2014non}, VIPL-HR~\cite{niu2018vipl}) are valuable resources. However, given the challenging nature of the rPPG task researchers have collected and released data under heavily constrained conditions with very little physical motion. More recent datasets (such as UBFC-PHYS~\cite{meziatisabour2021ubfc} and MMPD~\cite{tang2023mmpd}) contain larger and more natural motions. However, the baseline results on these datasets are not very strong. 

\vspace{-0.5em}
\section{Motion Augmented rPPG Video Pipeline}
\vspace{-0.5em}

We propose neural motion transfer as a data augmentation technique to train machine learning models for predicting physiological measurements, specifically photoplethysmography (PPG) signal, from facial videos. First, we describe our proposed pipeline to augment facial videos with naturalistic human head motion and expression in \Cref{sec:method_motion_transfer}. Neural motion transfer algorithms often use generative models to synthesize new videos of a person by transferring the rigid head motion and non-rigid facial expressions from a driving video of another person. Since these models generate image pixels from scratch, it is possible that images generated by neural motion transfer algorithms can destroy the underlying physiological signal. Thus, in \Cref{sec:evidence}, we provide qualitative evidence to prove that neural motion transfer algorithms do not destroy the original PPG signal, and the original heart rate is preserved. This allows us to effectively use neural motion transfer as a data augmentation technique for training rPPG networks. We provide additional quantitative evidence to highlight preservation of the underlying physiological signal through the signal-to-noise ratio (SNR) metric and after rPPG signal extraction using TS-CAN in Table~\ref{tab:compare_motion_transfer}. 

\begin{figure*}[ht]
    \begin{center}
    \includegraphics[width=\textwidth]{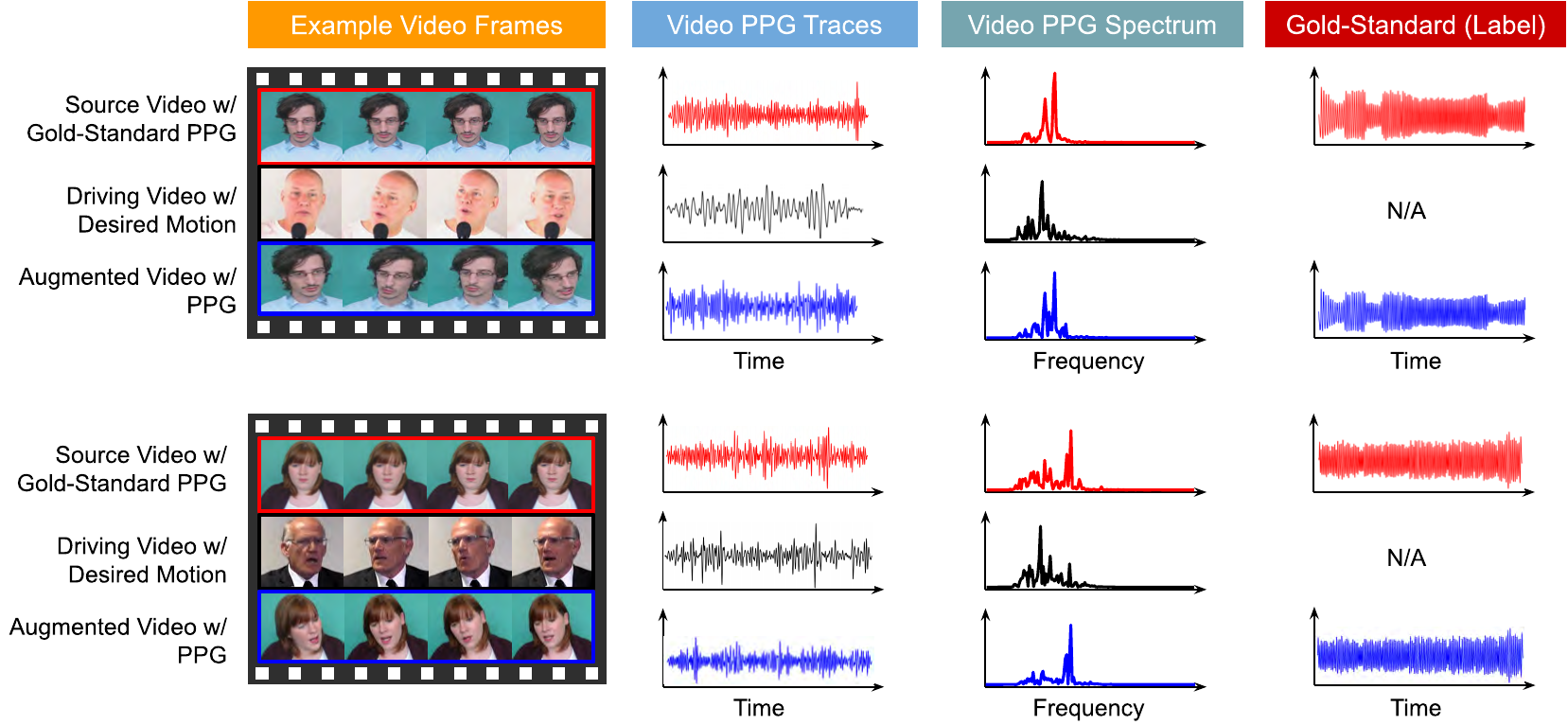}
    \end{center}
    \vspace{-0.5cm}
    \caption{\small{\textbf{Preserving physiological signals in motion augmented videos.} We show that applying neural motion transfer preserve the physiological signal corresponding to the heart-rate present in the peak of the frequency spectrum of the source and the augmented video.}}
    \label{fig:example_ma}
    \vspace{-0.35cm}
    \end{figure*}

\subsection{Motion Augmentation Pipeline}
\label{sec:method_motion_transfer}

In a camera-based physiological sensing (e.g., rPPG) task, a machine learning model is trained on facial videos with time-aligned physiological labels. These may take the form of continuous waveforms (e.g., a gold-standard PPG or a respiration wave) or vital statistics (e.g., heart or breathing rates). In this project, we consider video labels in the form of a PPG signal. The goal of designing a data augmentation strategy is to apply more naturalistic motion to the facial videos without changing the PPG labels.

To apply naturalistic motion to these facial videos, we consider neural talking-head video synthesis models that transfer more naturalistic motion from a \emph{driving} video of a person to the \emph{source} video associated with a PPG signal label. Our goal is to find a neural motion transfer algorithm that can: (a) inject a large variety of rigid and non-rigid head motions into the source video, (b) not introduce any artifacts that significantly degrade the generated video quality, and (c) maintain the key properties of the underlying PPG signal in terms of frequency information indicating physiological signals like heart rate.

Our pipeline takes in a source video with a PPG signal label from the training data, $\textbf{S}$, and a driving video, $\textbf{D}$, randomly selected from a curated driving video set as inputs for motion augmentation. Both $\textbf{S}$ and $\textbf{D}$ can be represented as a sequence of frames, respectively $\{s_1, s_2,...,s_n\}$ and $\{d_1, d_2,...,d_n\}$. Motion is transferred from driving video $\textbf{D}$ to source video $\textbf{S}$ on a frame-by-frame basis, such that an output video $\textbf{Y}$ represents the motion-augmented sequence of frames $\{y_1, y_2,...,y_n\}$. Thus we search for a motion transfer algorithm $M(\cdot;\theta)$, such that $y_t = \textbf{M}(s_t, d_t;\theta)$.

We choose Face-Vid2Vid~\cite{Wang2020OneShotFN}, a neural talking-head synthesis model for transferring motion from a driving video to a source video. The original Face-Vid2Vid paper was intended for teleconferencing applications where a motion-augmented video is generated from a single source image using a driving video. In contrast, we re-purpose the same core algorithm such that each frame of the source video is augmented with motion from the corresponding frame of the driving video. The motion-augmented video $\textbf{Y}$, along with the original PPG signal label, is ultimately used as training data for various deep learning-based camera physiological measurements. This pipeline is shown in \Cref{fig:pipeline} and is further described in our supplementary materials, alongside provided code that will be open-sourced.

\noindent\textbf{Source Video Datasets:} We utilize the UBFC-rPPG~\cite{bobbia2019unsupervised} and PURE~\cite{stricker2014non} rPPG video datasets as source videos. The UBFC-rPPG dataset contains videos with a very minimal amount of both rigid motion and non-rigid motion, making them ideal for motion augmentation. The PURE dataset contains videos of various tasks with a variety of constrained rigid and non-rigid motion.

\noindent\textbf{Driving Video Datasets:} The driving video datasets used include a self-captured, constrained driving video set (CDVS) and the TalkingHead-1KH~\cite{Wang2020OneShotFN} dataset. The CDVS contains 90 self-captured videos by 5 subjects with heavily constrained, unnatural motion used only for ablation studies to understand the impact of augmenting data with various degrees of rigid and non-rigid motion. The CDVS will be released in the future for research purposes. Talkinghead-1KH is a publicly available, large-scale talking-head video dataset used as a benchmark for Face-Vid2Vid~\cite{Wang2020OneShotFN} and entirely sourced from YouTube videos. It contains 180K unconstrained videos of people speaking in a variety of real-world contexts, leading to a rich diversity in both rigid and non-rigid motion.

\noindent\textbf{Deep Networks for estimating PPG signal:} For our experiments, we focus on using TS-CAN~\cite{liu2020multi} to predict the 1st-order derivative of the PPG signal after training on videos augmented with motion. We also use DeepPhys~\cite{chen2018deepphys} and PhysNet~\cite{yu2019remote} to highlight the consistent benefits of motion augmentation across different neural models.

\subsection{The Effect of Motion Transfer on PPG}
\label{sec:evidence}

Neural Motion Transfer algorithms are based on generative models where every pixel of the generated image is synthesized by a neural network. While these algorithms succeed in producing photorealistic facial images that are indistinguishable from real images, it is not obvious if the synthesized videos can preserve the underlying PPG signal.

In an ideal world, a motion transfer algorithm is expected to perturb the PPG signal since head motion will induce certain changes in raw pixel intensities. However, the frequency domain analysis of the PPG signal should preserve the peaks related to the heart rate of the patient. It is highly unlikely that the peak frequency of head motion and heart rate will be exactly the same.

Thus, our goal is to first analyze if the motion transfer algorithm of Face-Vid2Vid~\cite{Wang2020OneShotFN} can preserve the peak heart rate indicated in the frequency domain analysis of the PPG signal extracted from the source video and the synthesized video. In \Cref{fig:example_ma}, we qualitatively analyze the time-domain and frequency domain PPG signals extracted from the source and the synthesized (augmented) video. We choose a simple unsupervised algorithm, POS~\cite{wang2017algorithmic}, for extracting the PPG signal from all the facial videos to focus more on the original signal contents in the videos. We observe that the most prominent frequency peak, corresponding to the heart rate, is the same for the source video and the augmented video. This appears to also hold true across different appearances and motion conditions, both in the source videos and the driving videos. Again, we also provide quantitative evidence to support our observation in \Cref{tab:compare_motion_transfer}. We also present additional qualitative results and information about prior works analyzing the presence of physiological signals in deep fake videos in the supplementary material. Thus, we can effectively claim that keypoint-based motion transfer algorithms like Face-Vid2Vid~\cite{Wang2020OneShotFN} do preserve the underlying physiological signal, like heart rate, and they can be a very effective tool for large scale augmentation of training videos for rPPG estimation tasks. Our quantitative experimental results show that deep neural networks for camera physiological measurement can take advantage of this to significantly improve model performance by training on motion-augmented data.

\vspace{-0.5em}
\section{Experiments}
\vspace{-0.5em}
\label{sec:experiments}

\begin{table}[b!]
    \small
        \captionsetup{width=\textwidth}
        \vspace{-2em}
        \captionsetup{width=\columnwidth}
        \caption{\small{\textbf{Comparison to SOTA for PURE dataset.} We compare our approach to other SOTA methods using the same source data, the UBFC-rPPG dataset. The best result is shown in bold.}}
        \vspace{-8pt}
        \label{tab:sota_pure}
        \centering
        \small{
        \setlength{\tabcolsep}{8pt}
        \adjustbox{max width=\columnwidth}{
        \begin{tabular}{llcc}
        \toprule[1.5pt]
        \textbf{Method} & MAE$\downarrow$ \\ \midrule
        EfficientPhys-C~\cite{Liu_2023_WACV} & 5.47 \\ 
        SiNC~\cite{Speth_2023_CVPR} & 4.02 \\ 
        PhysNet~\cite{yu2019remote} & 3.81 \\ 
        Physformer~\cite{yu2021physformer} & 1.99 \\ 
        Dual-GAN~\cite{lu2021dual} & 1.81 \\ 
        \grayrow
        Ours (Motion Augmented) & \textbf{0.96} \\ 
        \midrule
        \multicolumn{1}{c}{\textsc{Ours vs. Best Baseline}} & \textbf{\textcolor{ForestGreen}{+46.96\%}} \\
        \bottomrule[1.5pt]
        \end{tabular}}
        \\
        }
        \scriptsize
        MAE = Mean Absolute Error in HR estimation (Beats/Min)
        \vspace{-1em}
    \vspace{-0.2cm}
\end{table}

We consider five datasets for training and evaluation, \textbf{UBFC-rPPG}~\cite{bobbia2019unsupervised}, \textbf{PURE}~\cite{stricker2014non}, \textbf{UBFC-PHYS}~\cite{meziatisabour2021ubfc}, \textbf{AFRL}~\cite{estepp2014recovering}, and \textbf{MMPD}~\cite{tang2023mmpd} (see supplementary materials for more details). They consist of facial videos and corresponding gold-standard PPG signal labels. We use some of these datasets for augmentation with neural motion transfer and training the rPPG models, and use the rest to evaluate different aspects of the effectiveness of neural motion augmentation. To our knowledge, we perform the most extensive inter-dataset evaluation of rPPG estimation to date, testing on five independent test datasets.

\begin{table*}[ht!]
    \small
        \captionsetup{width=\textwidth}
        \caption{\small{\textbf{Evaluation across all datasets.} We motion-augment two training datasets, UBFC-rPPG and PURE, to create MAUBFC-rPPG and MAPURE, respectively. We observe that the motion-augmented versions produce significant improvements (shown in bold).}}
        \vspace{-8pt}
        \label{tab:main_overall_results}
        \centering
        \small
        \setlength{\tabcolsep}{4pt}
        \adjustbox{max width=\textwidth}{
        \begin{tabular}{llcccccccccc}
        \toprule[1.5pt]
            & & \multicolumn{10}{c}{\textbf{Testing Set}} \\
            & & \multicolumn{2}{c}{\textbf{UBFC-rPPG}} & \multicolumn{2}{c}{\textbf{PURE}} & \multicolumn{2}{c}{\textbf{UBFC-PHYS}} & \multicolumn{2}{c}{\textbf{AFRL}} & \multicolumn{2}{c}{\textbf{MMPD}} \\
            \cmidrule(lr){3-4} \cmidrule(lr){5-6} \cmidrule(lr){7-8} \cmidrule(lr){9-10} \cmidrule(lr){11-12}
        \textbf{Training Set} & \multicolumn{1}{c}{\textbf{Method}} & MAE$\downarrow$ &  MAPE$\downarrow$ & MAE$\downarrow$ &  MAPE$\downarrow$ & MAE$\downarrow$ &  MAPE$\downarrow$ & MAE$\downarrow$ &  MAPE$\downarrow$ & MAE$\downarrow$ &  MAPE$\downarrow$ \\ \midrule\midrule
        \multirow{4}{*}{Unsupervised} & \multicolumn{1}{c}{Green} & 19.82 & 18.78 & 10.09 & 10.28 & 13.45 & 16.00 & 7.01 & 9.24 & 16.27 & 20.09 \\
         & \multicolumn{1}{c}{ICA} & 14.70 & 14.34 & 4.77 & 4.47 & 8.00 & 9.48 & 6.77 & 8.96 & 13.10 & 16.33 \\
        & \multicolumn{1}{c}{CHROM} & 3.98 & 3.78 & 5.77 & 11.52 & 4.68 & 6.20 & 5.41 & 7.95 & 8.85 & 11.93 \\
         & \multicolumn{1}{c}{POS} & 4.00 & 3.86 & 3.67 & 7.25 & 4.62 & 6.29 & 6.93 & 10.00 & 8.18 & 11.12 \\ \midrule
        UBFC-rPPG & \multicolumn{1}{c}{TS-CAN} & - & - & 4.55 & 4.67 & 5.56 & 7.25 & 4.24 & 5.84 & 8.74 & 10.51 \\
        \grayrow
        MAUBFC-rPPG & \multicolumn{1}{c}{TS-CAN} & - & - & \textbf{0.96} & \textbf{1.13} & \textbf{3.93} & \textbf{5.24} & 2.67 & 3.65 & \textbf{6.80} & \textbf{7.97} \\
        PURE & \multicolumn{1}{c}{TS-CAN} & 1.34 & 1.55 & - & - & 4.43 & 5.89 & 2.63 & 3.51 & 8.96 & 10.33 \\
        \grayrow
        MAPURE & \multicolumn{1}{c}{TS-CAN} & \textbf{1.03} & \textbf{1.17} & - & - & 4.39 & 5.90 & \textbf{2.37} & \textbf{3.26} & 8.08 & 9.54 \\
        \midrule
        \multicolumn{2}{c}{\textsc{MAUBFC-rPPG vs. UBFC-rPPG}} & - & - & \textbf{\textcolor{ForestGreen}{+78.90\%}} & \textbf{\textcolor{ForestGreen}{+75.08\%}} & \textbf{\textcolor{ForestGreen}{+29.32\%}} & \textbf{\textcolor{ForestGreen}{+27.72\%}} & \textbf{\textcolor{ForestGreen}{+37.03\%}} & \textbf{\textcolor{ForestGreen}{+37.50\%}} & \textbf{\textcolor{ForestGreen}{+22.20\%}} & \textbf{\textcolor{ForestGreen}{+24.17\%}} \\
        \midrule
        \multicolumn{2}{c}{\textsc{MAPURE vs. PURE}} & \textbf{\textcolor{ForestGreen}{+23.13\%}} & \textbf{\textcolor{ForestGreen}{+24.52\%}} & - & - & \textbf{\textcolor{ForestGreen}{+0.90\%}} & \textbf{\textcolor{darkred}{-0.17\%}} & \textbf{\textcolor{ForestGreen}{+9.89\%}} & \textbf{\textcolor{ForestGreen}{+7.12\%}} & \textbf{\textcolor{ForestGreen}{+9.82\%}} & \textbf{\textcolor{ForestGreen}{+7.65\%}} \\

        \bottomrule[1.5pt]
        \end{tabular}}
        \footnotesize
       MAE = Mean Absolute Error in HR estimation (Beats/Min), MAPE = Mean Absolute Percentage Error in HR estimation
    \vspace{-0.2cm}
\end{table*}

\textbf{Implementation Details:} The predicted PPG signals were filtered using a band-pass filter with cut-offs 0.75 Hz and 2.5 Hz. The heart rate was calculated based on the predicted PPG signal using the Fast Fourier Transform (FFT), with a measurement window of the video length.
All networks were trained using an NVIDIA RTX A4500 and PyTorch~\cite{paszke2019pytorch} implementations in a publicly available toolbox for the rPPG task~\cite{liu2022rppg}. A cyclic learning rate scheduler was utilized with 30 epochs, a learning rate of 0.009, and a batch size of 4 for both training and inference.

\subsection{Training with Motion Augmented Data}
\label{sec:trainingma}

In \Cref{tab:sota_pure}, we compare our approach with TS-CAN and motion-augmented source data to other SOTA methods using the same source data, the UBFC-rPPG dataset. The PhysNet~\cite{yu2019remote} result was obtained from~\cite{Speth_2023_CVPR}, and differs from our reproduced PhysNet result in \Cref{tab:main_neural_methods_ablation} due to pre-processing and implementation differences. We achieve a 47\% improvement over SOTA results on the PURE dataset with our data augmentation strategy using neural motion transfer. In \Cref{tab:main_overall_results}, we comprehensively compare the performance of a supervised PPG estimation network, TS-CAN~\cite{liu2020multi}, trained on existing video datasets and motion-augmented versions of those datasets. We also show the performance of unsupervised methods for comparison. For the sake of space and clarity, the aforementioned tables only show limited metrics such as the MAE or MAPE in heartrate estimation. Equivalent tables with additional metrics, including root mean squared error (RMSE) and Pearson correlation metrics can be found where applicable in the supplementary material. The driving videos used for augmentation in \Cref{tab:main_overall_results} contain significant amounts of unconstrained motion -- both rigid and non-rigid.

We observe that training TS-CAN on augmented videos produces SOTA performance in most cases. Additionally, we observe that in most cases, the augmented versions outperform the non-augmented versions, with a gain in performance up to 79\% and an average gain of 26\%. However, when comparing the performance of MAPURE versus PURE when tested on UBFC-PHYS, we note a minor drop in performance rather than an improvement due to the difficulty in effectively augmenting the PURE dataset. This is because the PURE dataset already contains significant amounts of rigid motion, and when augmented, it may provide training data with artifacts that make the learned rPPG task less useful in the face of a highly unconstrained dataset with natural rigid and non-rigid motion.

\textbf{Details:} We utilize all downloadable videos from the TalkingHead-1KH~\cite{Wang2020OneShotFN} dataset as our driving videos for augmenting various rPPG video datasets with motion. We analyze the videos using OpenFace~\cite{baltruvsaitis2016openface} to obtain the intensity (0 to 5) of 17 Facial Action Units (AUs) and the head pose rotations $R_x$, $R_y$, and $R_z$ in radians (rad). To generate MAUBFC-rPPG, we choose driving videos from a pool of 60 driving videos with a range of mean standard deviation in head pose rotations from 0.10 to 0.14 rad to augment as much rigid motion as possible into a source video dataset that has very little of both rigid and non-rigid motion. We do not constrain for non-rigid motion in this case, so we observe a wide range of mean standard deviation in facial AUs from 0.15 to 0.5 intensity. To generate MAPURE, we choose driving videos with a range of mean standard deviation in facial AUs from 0.45 to 0.55 intensity to augment as much non-rigid motion as possible into a source video dataset that has very little non-rigid motion. We do not constrain for rigid motion in this case, so we observe a wide range of mean standard deviation in head pose rotations from 0.03 to 0.14 rad.

\subsection{Effect of Motion Types}
\label{sec:m_type}

A key question in designing a motion augmentation strategy is deciding what type of motion should be applied to obtain the best performance on a certain evaluation dataset. To answer this question, we separately analyze two types of motion: rigid and non-rigid, by augmenting training data with different magnitudes of motion. Rigid motion refers to head pose rotation, while having minimal change in facial action units or expressions. Non-rigid motion refers to changes in facial expression, i.e. motion in facial action units for various tasks like talking, while having minimal head pose rotation.

\textbf{Rigid Motion:} For rigid motion, we consider UBFC-rPPG as training data, which has very little head motion and AFRL as test data which has large variations in rigid head motion. We classify videos in the AFRL dataset into different rigid head motion categories: 'very small motion', 'small motion' (10 deg rotation per sec), and 'large motion' (30 deg rotation per sec). Based on this categorization, we also select driving videos from our captured CDVS to have 'small motion' and 'large motion' using the mean standard deviation in estimated head pose rotations across all the frames of a video. Specifically, for 'small motion' we used mean standard deviation between 0.03 to 0.07 rad and for 'large motion' between 0.10 to 0.14 rad. These parameters are chosen to roughly match the distribution of head pose rotation in 'small motion' and 'large motion' categories of AFRL. We then use these videos from the CDVS dataset to augment the source videos of UBFC-rPPG to create 3 separate categories of augmented videos for 'very small motion' (which is the original UBFC-rPPG dataset), 'small motion', and 'large motion' respectively. We then train TS-CAN on augmented data in each category and test on the same categories of the AFRL dataset. We present these results in \Cref{tab:main_rigid}.

We observe that when the test data of AFRL has 'very small motion' or 'small motion', augmenting UBFC-rPPG with small motion performs the best. In fact, augmenting with large motion worsens the result by 19\% in this case. However, when testing on the 'large motion' split of AFRL, UBFC-rPPG augmented with 'large motion' outperforms 'small motion' by 13.5\% and 'very small motion' by 52\%.

\textbf{Non-rigid Motion:} For non-rigid motion, we also consider UBFC-rPPG as training data since it has very little motion, and the speech task of the PURE dataset~\cite{stricker2014non} as the test data which has significant non-rigid head motion. We also augment the UBFC-rPPG dataset with non-rigid head motion from our captured CDVS with 'small' and 'large' non-rigid motions and minimal rigid motion. For this experiment, we define small non-rigid motion to have a range of mean standard deviation in facial action units from 0.15 to 0.25 intensity and large non-rigid motion to have a range of mean standard deviation in facial action units from 0.45 to 0.55 intensity. We train TS-CAN on 'small' and 'large' motion augmented versions of UBFC-rPPG and test it on the speech task of PURE, in which recorded participants are asked to talk while avoiding head movements as much as possible. We present these results in \Cref{tab:main_nonrigid}.
We observe that augmenting UBFC-rPPG with 'large' non-rigid motion improves over 'very small motion' (original UBFC-rPPG) by 89.2\% and over 'small' non-rigid motion by 37\%.

\begin{table}[ht!]
    \vspace{-1em}
    \small
        \captionsetup{width=\columnwidth}
        \caption{\small{\textbf{Effect of Motion Types -- Rigid.} We augment UBFC-rPPG with various types of rigid head motions and test on AFRL~\cite{estepp2014recovering}. The best results are shown in bold.}}
        \vspace{-8pt}
        \label{tab:main_rigid}
        \centering
        \small
        \setlength{\tabcolsep}{1pt}
        \adjustbox{max width=\columnwidth}{
        \begin{tabular}{llcccc}
        \toprule[1.5pt]
            & & \multicolumn{4}{c}{\textbf{Testing Set}} \\
            & & \multicolumn{1}{c}{\textbf{No}} & \multicolumn{1}{c}{\textbf{Small}} & \multicolumn{1}{c}{\textbf{Large}} & \multicolumn{1}{c}{\textbf{All}} \\
            & & \multicolumn{1}{c}{\textbf{Motion}} & \multicolumn{1}{c}{\textbf{Motion}} & \multicolumn{1}{c}{\textbf{Motion}} & \multicolumn{1}{c}{\textbf{Motion}} \\
            \cmidrule(lr){3-3} \cmidrule(lr){4-4} \cmidrule(lr){5-5} \cmidrule(lr){6-6}
        \textbf{Training Set} & \textbf{Rigid Motion} & MAE$\downarrow$ & MAE$\downarrow$ & MAE$\downarrow$ & MAE$\downarrow$ \\ \midrule\midrule
        UBFC-rPPG & \multicolumn{1}{c}{Very Small} & 1.00 & 2.28 & 7.59 & 4.72 \\
        \grayrow
        MAUBFC-rPPG & \multicolumn{1}{c}{Small} & \textbf{0.84} & \textbf{1.44} & 4.21 & \textbf{3.19} \\
        \grayrow
        MAUBFC-rPPG & \multicolumn{1}{c}{Large} & 1.00 & 1.78 & \textbf{3.64} & 3.39 \\
        \midrule
        \multicolumn{2}{c}{\textsc{Ours vs. Baseline}} & \textbf{\textcolor{ForestGreen}{+16.0\%}} & \textbf{\textcolor{ForestGreen}{+36.8\%}} & \textbf{\textcolor{ForestGreen}{+52.0\%}} & \textbf{\textcolor{ForestGreen}{+32.4\%}} \\
        \bottomrule[1.5pt]
        \end{tabular}}
    \vspace{-0.2cm}
\end{table}

\begin{table}[ht!]
    \vspace{-1em}
    \small
        \captionsetup{width=\columnwidth}
        \caption{\small{\textbf{Effect of Motion Types -- Non-rigid.} We augment UBFC-rPPG with various types of non-rigid motions (expressions) and test on the speech task, in PURE~\cite{stricker2014non}. The best results are shown in bold.}}
        \vspace{-8pt}
        \label{tab:main_nonrigid}
        \centering
        \small
        \setlength{\tabcolsep}{4pt}
        \adjustbox{max width=\columnwidth}{
        \begin{tabular}{llcc}
        \toprule[1.5pt]
            & & \multicolumn{2}{c}{\textbf{Testing Set}} \\
            & & \multicolumn{2}{c}{\textbf{Non-rigid Motion Task}} \\
            \cmidrule(lr){3-4}
        \textbf{Training Set} & \textbf{Non-Rigid Motion} & MAE$\downarrow$ &  MAPE$\downarrow$ \\ \midrule\midrule
        UBFC-rPPG & \multicolumn{1}{c}{Very Small} & 10.84 & 11.40 \\
        \grayrow
        MAUBFC-rPPG & \multicolumn{1}{c}{Small} & 1.86 & 2.94 \\
        \grayrow
        MAUBFC-rPPG & \multicolumn{1}{c}{Large} & \textbf{1.17} & \textbf{1.55} \\
        \midrule
        \multicolumn{2}{c}{\textsc{Ours vs. Baseline}} & \textbf{\textcolor{ForestGreen}{+89.2\%}} & \textbf{\textcolor{ForestGreen}{+86.4\%}} \\
        \bottomrule[1.5pt]
        \end{tabular}}
    \vspace{-0.4cm}
\end{table}

\subsection{Naturalistic versus Synthetic Head Motion}
\label{sec:m_graphics}

\begin{table}[b!]
    \vspace{-0.6cm}
    \small
        \captionsetup{width=\columnwidth}
        \caption{\small{\textbf{Naturalistic versus Synthetic Head Motion.} We evaluate the effect of adding head motions to SCAMPS and UBFC-rPPG. The best results are shown in bold.}}
        
        \vspace{-8pt}
        \label{tab:main_synthetic_versus_real}
        \centering
        \small
        \setlength{\tabcolsep}{8pt}
        \adjustbox{max width=\columnwidth}{
        \begin{tabular}{lccccc}
        \toprule[1.5pt]
            & \multicolumn{4}{c}{\textbf{Testing Set}} & \\
            & \multicolumn{2}{c}{\textbf{PURE}} & \multicolumn{2}{c}{\textbf{AFRL}} \\
            \cmidrule(lr){2-3} \cmidrule(lr){4-5}
        \textbf{Training Set} & MAE$\downarrow$ &  MAPE$\downarrow$ & MAE$\downarrow$ &  MAPE$\downarrow$ & \multicolumn{1}{c}{Synth. Time} \\ \midrule\midrule
        SCAMPS-200 (No motion) & 10.29 & 11.09 & 7.75 & 10.54 & 37.00s \\
        SCAMPS-200 (Motion) & 5.38 & 5.42 & 7.25 & 10.20 & 37.00s \\
        UBFC-rPPG & 4.55 & 4.67 & 4.72 & 6.59 & - \\
        \grayrow
        MASCAMPS-200 & 4.67 & 4.22 & 5.00 & 6.69 & 1.20s \\
        \grayrow
        MAUBFC-rPPG & \textbf{0.96} & \textbf{1.13} & \textbf{3.24} & \textbf{4.37} & 2.39s \\
        \midrule
        \multicolumn{1}{c}{\textsc{MASCAMPS vs. SCAMPS}} & \textbf{\textcolor{ForestGreen}{+13.2\%}} & \textbf{\textcolor{ForestGreen}{+22.1\%}} & \textbf{\textcolor{ForestGreen}{+31.1\%}} & \textbf{\textcolor{ForestGreen}{+34.4\%}} & \textbf{\textcolor{ForestGreen}{+96.8\%}} \\
        \midrule
        \multicolumn{1}{c}{\textsc{MAUBFC vs. UBFC-rPPG}} & \textbf{\textcolor{ForestGreen}{+78.9\%}} & \textbf{\textcolor{ForestGreen}{+75.8\%}} & \textbf{\textcolor{ForestGreen}{+31.4\%}} & \textbf{\textcolor{ForestGreen}{+33.7\%}} & - \\
        \bottomrule[1.5pt]
        \end{tabular}}
        \footnotesize{
       Avg. Synth. Time = time (in seconds) to synthesize a frame}
    \vspace{-0.3cm}
\end{table}

In order to further evaluate the impact of motion transfer as a data augmentation technique, we explore whether data augmented with natural head motion using a neural motion transfer algorithm is better than synthetic data with motion generated using parametric motion animation, as used in the SCAMPS dataset~\cite{mcduff2022scamps}. The SCAMPS dataset consists of synthetic human heads that can be rigged to induce parametric motion. We consider 200 such samples from the SCAMPS dataset that consist of significant synthetically generated rigid and non-rigid head motion. We then take instances from the SCAMPS dataset with no head motion and augment them with naturalistic head motion to produce MASCAMPS-200. Further details on this experiment, including comparisons to additional results using Wang et al.'s~\cite{wang2022synthetic} synthetic rPPG video data, are included in the supplementary materials.

We train TS-CAN on both SCAMPS-200 (Motion) and MASCAMPS-200, and evaluated its performance on PURE and AFRL, as shown in \Cref{tab:main_synthetic_versus_real}. We observed that adding naturalistic motion improved performance by 13.2\% on PURE and 31.1\% on AFRL compared to synthetically generated motion. It is worth noting that the average time taken to add synthetic motion to each frame of a sequence is 37 seconds, compared to only 1.2 seconds for adding naturalistic motion using the neural motion transfer algorithm. For comparison, we also included real-world training data, UBFC-rPPG, which showed that having real images significantly improved performance over synthetic images. Furthermore, the only way to augment real images is to use the neural motion transfer algorithm, as parametric rigged head motion cannot be applied to real data.

\vspace{-0.2cm}

\subsection{Effect of Neural Motion Transfer Algorithms}
\label{sec:effect_of_motion_transfer_algorithms}

It is important to decouple any data augmentation technique from additional factors that affect its usefulness for a given set of training data. One such factor is the neural motion transfer algorithm used for motion augmentation. In \Cref{tab:compare_motion_transfer}, we evaluate two additional neural motion transfer methods in addition to face-vid2vid~\cite{Wang2020OneShotFN} - FOMM~\cite{Siarohin_2019_NeurIPS} and DaGAN~\cite{Hong2022DepthAwareGA}. MAE and SNR are calculated using predictions from TS-CAN and the ground truth label. These results show that most motion transfer algorithms can serve as an effective data augmentation tool as long as they utilize a keypoint-based approach for transferring motion toward applications such as neural talking head synthesis.

\begin{table}[h!]
    \small
        \captionsetup{width=\columnwidth}
        \vspace{-1em}\caption{\small{\textbf{Effect of Motion Transfer Methods.} We compare numerous SOTA neural motion transfer methods on UBFC-rPPG~\cite{bobbia2019unsupervised}.}}
        \vspace{-8pt}
        \label{tab:compare_motion_transfer}
        \centering
        \small
        \setlength{\tabcolsep}{8pt}
        \adjustbox{max width=\columnwidth}{
        \begin{tabular}{lcc}
        \toprule[1.5pt]
        \textbf{Method} & MAE$\downarrow$ & SNR $\uparrow$ \\ \midrule\midrule
        Baseline (No Augmentation) & 3.93 & 4.72 \\
        FOMM~\cite{Siarohin_2019_NeurIPS} & 0.92 & 8.64 \\
        DaGAN~\cite{Hong2022DepthAwareGA} & 1.23 & 8.37 \\
        \grayrow
        face-vid2vid~\cite{Wang2020OneShotFN} & 0.96 & 8.70 \\
        \bottomrule[1.5pt]
        \end{tabular}}
        \\
        \vspace{-0.6cm}
\end{table}

\subsection{Effect of rPPG Estimation Models}
\label{m_consistency}

It is important to decouple any data augmentation technique from additional factors that affect its usefulness for a given set of training data. One such factor is the neural network model used for training and evaluation. Thus, in addition to TS-CAN, we evaluate two more rPPG models - DeepPhys and PhysNet - in \Cref{tab:main_neural_methods_ablation}. We utilize MAUBFC-rPPG as training data and evaluate on PURE. We observe that the results are reasonably consistent across neural rPPG models.

\begin{table}[h!]
    \vspace{-1em}
    \small
        \captionsetup{width=\columnwidth}
        \caption{\small{\textbf{Generalization to Different rPPG Models.} We train different PPG estimation networks on UBFC-rPPG and MAUBFC-rPPG and evaluate on PURE. The best results are shown in bold.}}
        
        \vspace{-8pt}
        \label{tab:main_neural_methods_ablation}
        \centering
        \small
        \setlength{\tabcolsep}{8pt}
        \adjustbox{max width=\columnwidth}{
        \begin{tabular}{llcc}
        \toprule[1.5pt]
            & & \multicolumn{2}{c}{\textbf{Testing Set}} \\
            & & \multicolumn{2}{c}{\textbf{PURE}} \\
            \cmidrule(lr){3-4}
        \textbf{Training Set} & \textbf{Method} & MAE$\downarrow$ & MAPE$\downarrow$ \\ \midrule\midrule
        UBFC-rPPG & DeepPhys \cite{chen2018deepphys} & 5.14 & 4.90 \\
        \grayrow
        MAUBFC-rPPG & DeepPhys & 1.24 & 1.56 \\
        UBFC-rPPG & PhysNet \cite{yu2019remote} & 8.06 & 13.67 \\
        \grayrow
        MAUBFC-rPPG & PhysNet & 2.38 & 2.44 \\
        UBFC-rPPG & TS-CAN \cite{liu2020multi} & 4.55 & 4.67 \\
        \grayrow
        MAUBFC-rPPG & TS-CAN & \textbf{0.96} & \textbf{1.13} \\
        \bottomrule[1.5pt]
        \end{tabular}}
    \vspace{-0.6cm}
\end{table}

\vspace{-0.6em}
\section{Discussion}
\vspace{-0.5em}
\label{sec:disc}

\textbf{Can motion augmented videos achieve SOTA results?}

We conducted a set of systematic empirical validation studies that show that these videos can be used to effectively train rPPG models that generalize to independent benchmark datasets (see \Cref{tab:main_overall_results}). Cross-dataset experiments show a 23.1\% reduction in HR MAE on UBFC-rPPG when using the motion-augmented PURE datasets for training and a 79\% reduction in HR MAE on PURE when using the motion-augmented UBFC-rPPG dataset for training. Other than PURE, the largest gains were observed training on MAUBFC-rPPG and testing on videos with large rigid and/or non-rigid head motions (UBFC-PHYS: 29.32\%, AFRL: 37.03\% and MMPD: 22.20\% reduction in HR MAE). In \Cref{tab:sota_pure}, we also demonstrate, using UBFC-rPPG as a source dataset, the effectiveness of our method in contrast to other SOTA methods using the same source dataset to test on PURE.

\textbf{What type of motion is best to augment?} In learning tasks, designing training data that matches the distribution of the testing data is advantageous. Does augmenting motion in the training set that is similar to that in a testing set lead to optimal results? Our experiments show that this is the case for both rigid (see \Cref{tab:main_rigid}) and non-rigid (see \Cref{tab:main_nonrigid}) head motions. Furthermore, if the motions have a larger magnitude, then including larger magnitude motions in the training set empirically seems to have a benefit.

\textbf{Does the type of motion transfer algorithm or the type of PPG estimation model matter?}

It's important to decouple other factors that may significantly affect performance such as the neural motion transfer algorithm or the type of rPPG estimation model. As per \Cref{tab:compare_motion_transfer} and \Cref{tab:main_neural_methods_ablation}, it is clear that our motion augmentation strategy provides significant improvements 1) regardless of the 
core neural motion transfer algorithm utilized and 2) despite differences (e.g., 2DCNN versus 3DCNN) in neural rPPG estimation models. This in turn shows great promise in motion augmentation as a general data augmentation strategy for rPPG videos.

\textbf{Is natural motion augmentation best?} Finally, there are different methods for synthesizing motion in video data. SOTA synthetic datasets are generated using parametric computer graphics, but they require a large amount of computational resources. As a result, if the motions present in those datasets are sub-optimal, it is costly to remedy. Can motion augmentation add motions to these datasets "cheaply" and still obtain the performance benefits of graphics approaches? Our results in \Cref{tab:main_synthetic_versus_real} suggest that the motion in the SCAMPS dataset is sub-optimal when tested on PURE and AFRL. We were able to obtain a performance gain by using our simple motion augmentation.

\textbf{What are the limitations of our method?} There are several limitations that we would like to highlight. First, detecting artifacts in augmented videos is not always trivial, and we used motion driving videos without extreme motions to mitigate the chance of augmented videos with unnatural artifacts. We did not conduct an extensive investigation to determine if other physiological changes (e.g., respiration) that might be correlated with the PPG signal are preserved in the augmented videos. However, empirically we have shown that these data can be used to effectively train \emph{heart rate} estimation models. We did not thoroughly test whether the waveform dynamics, beyond the dominant frequency, were faithfully preserved in the augmented videos. For tasks such as blood pressure estimation from PPG waveforms, morphological information is important. Our method does not address diversity across other dimensions, particularly identity diversity. The augmented datasets we produced, while contributing to significant improvements over the baselines, only contain examples from the same number of subjects as the original dataset. Other synthetic generation techniques~\cite{wang2022synthetic} could help in these regards alongside more generic neural rendering approaches such as ours.

\vspace{-0.5em}
\section{Conclusion}
\vspace{-0.5em}

Motion artifacts are a significant challenge in camera physiological measurement. The PPG signal presents only very subtle changes in diffuse light reflections from the skin, whereas motion of the head causes large changes in specular reflections. We have shown that neural motion augmentation can be used to create training data with more motion, while still preserving the pulse signal. Motion augmented data leads to up to 79\% reduction in error in cross-dataset experiments using TS-CAN and a 47\% reduction in error when compared to other state-of-the-art methods using the same source dataset.


\clearpage

\appendix

\section{Overview of Appendices}

Our appendices contain the following additional details and results:

\begin{itemize} 
    \itemsep0em 
    \item Tables~\ref{tab:nonrigid},~\ref{tab:rigid},~\ref{tab:overall_results},~\ref{tab:synthetic_versus_real}, and~\ref{tab:neural_methods_ablation} in Section~\ref{sec:metrics_and_blandaltman} include additional metrics, RMSE and the Pearson correlation coefficient, for experimental results included in the main paper. We also provide Scatter and Bland-Altman Plots in Figures~\ref{fig:bland-altman-ubfc_maubfc} and~\ref{fig:bland-altman-pure_mapure} that correspond to the overall results shown in \Cref{tab:overall_results}. Section~\ref{sec:details} contains additional details regarding our experimental process. Section~\ref{sec:m_scale} contains an additional experiment toward the effect of scaling a dataset using motion augmentation.
    \item Section~\ref{sec:intradataset}, and the corresponding \Cref{tab:pure_intradataset}, describe and show intra-dataset results using the PURE~\cite{stricker2014non} dataset.
    \item Section~\ref{sec:motion_augmented_videos} briefly describes additional materials that we provide for research purposes, including our motion augmentation pipeline code, pre-trained models, and motion analysis scripts. Additionally, Section~\ref{sec:ma_effects} shows more qualitative examples of the effects of motion augmentation on the underlying PPG signal, as well as briefly addresses prior works involving analysis of physiological signals in deep fake videos.
    \item Sections~\ref{sec:source_datasets},~\ref{sec:driving_datasets}, and~\ref{sec:eval_datasets}, provide further details on source, driving, and evaluation datasets used in the main paper.
    \item Section~\ref{sec:broaderimpact} is our broader impact statement.
\end{itemize}

\section{Experimental Results}
\label{sec:metrics_and_blandaltman}

\begin{figure*}[h!]
    \begin{center}
    \includegraphics[width=\textwidth]{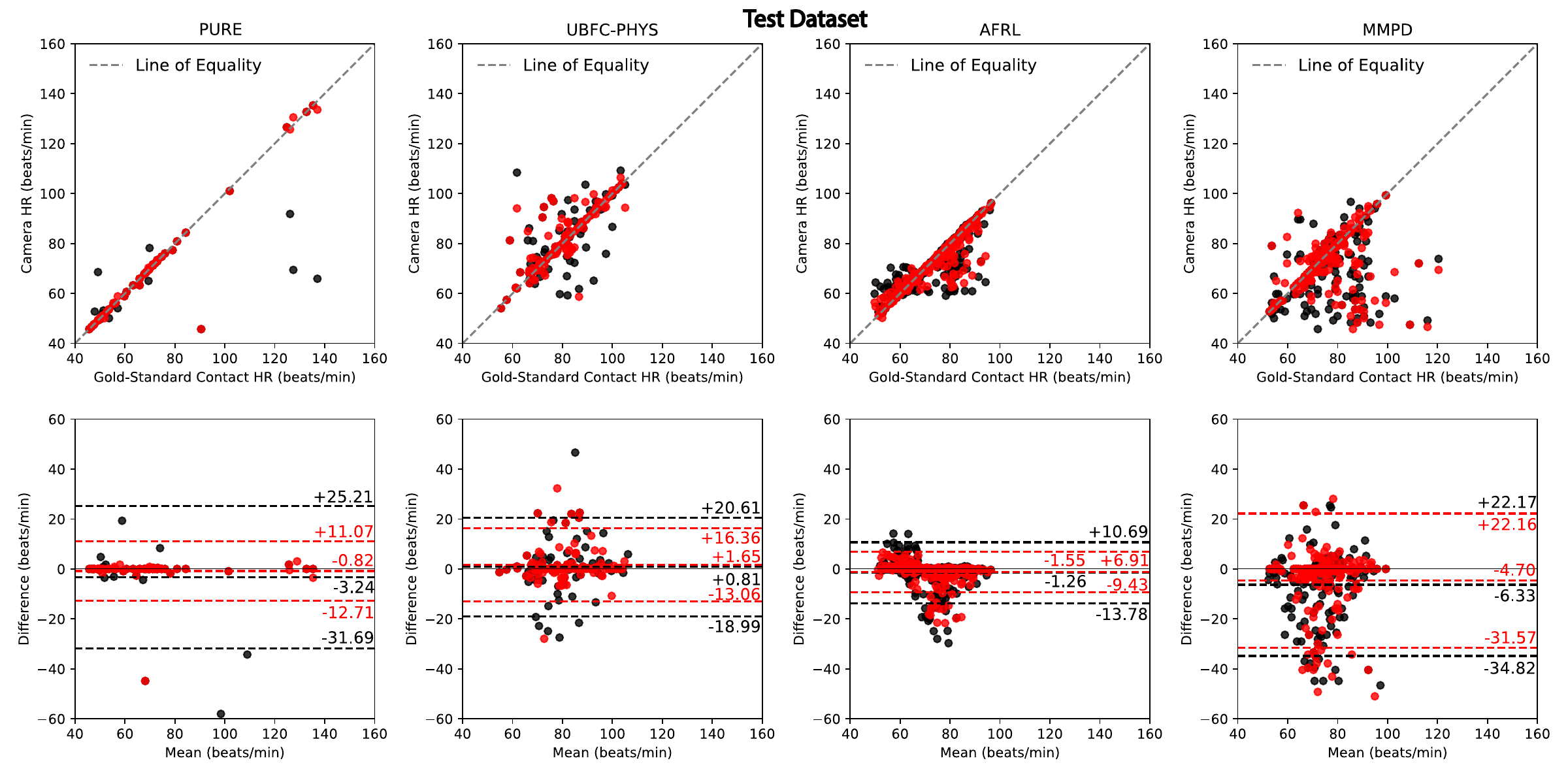}
    \end{center}
    \caption{\small{\textbf{Scatter and Bland-Altman Plots.} Scatter (top row) and Bland-Altman (bottom row) plots for models trained on UBFC-rPPG (black) and MAUBFC-rPPG (red) and tested on (from left to right), PURE, UBFC-PHYS, AFRL, and MMPD.}}
    \label{fig:bland-altman-ubfc_maubfc}
    \end{figure*}

\begin{figure*}[h!]
    \begin{center}
    \includegraphics[width=\textwidth]{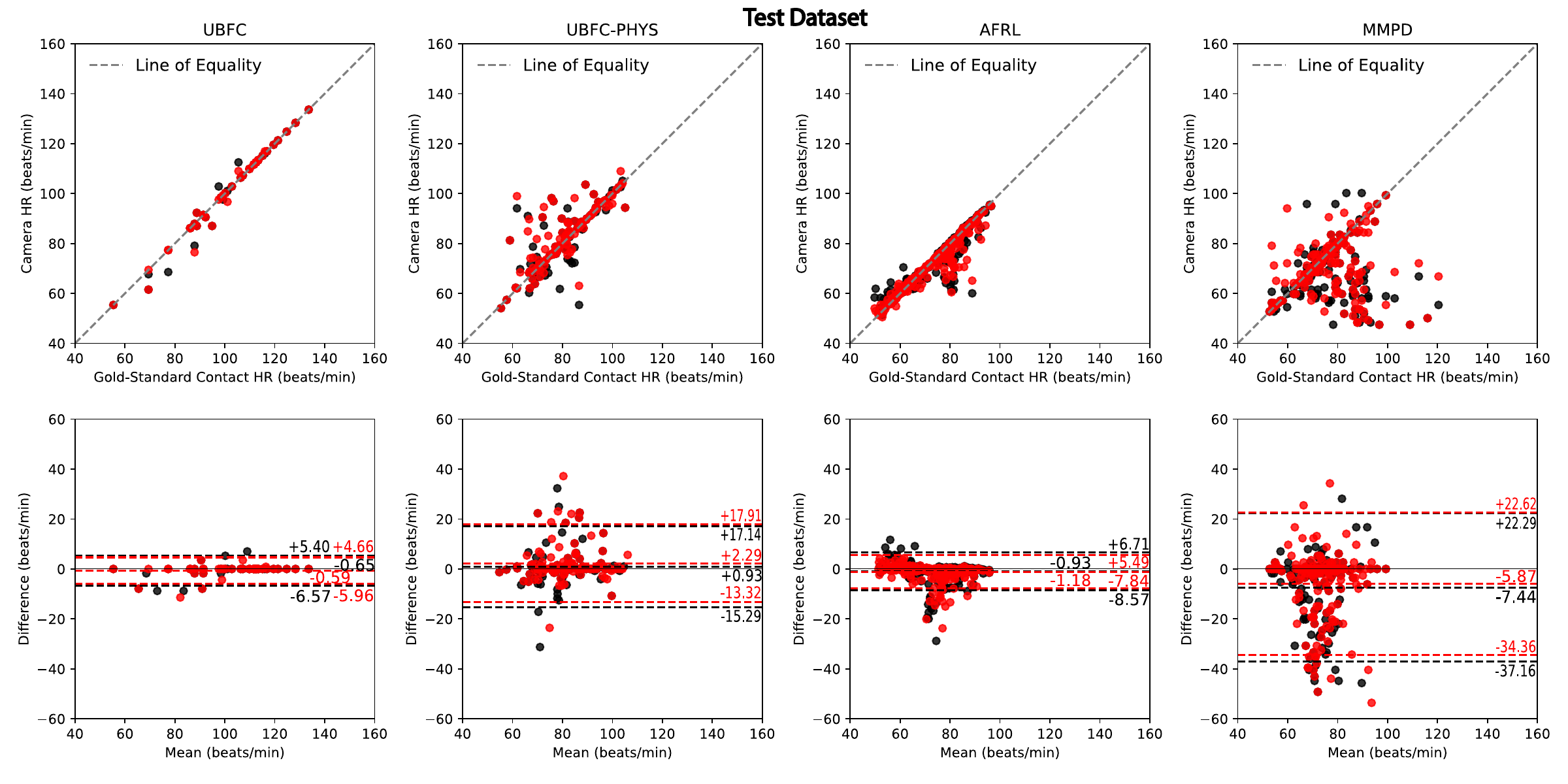}
    \end{center}
    \caption{\small{\textbf{Scatter and Bland-Altman Plots.} Scatter (top row) and Bland-Altman (bottom row) plots for models trained on PURE (black) and MAPURE (red) and tested on (from left to right), UBFC-rPPG, UBFC-PHYS, AFRL, and MMPD.}}
    \label{fig:bland-altman-pure_mapure}
    \vspace{1cm}
    \end{figure*}

The following section contains tables that include additional metrics, RMSE and the Pearson correlation coefficient, for experimental results already included in the main paper. We also provide scatter and Bland-Altman plots in Figures~\ref{fig:bland-altman-ubfc_maubfc} and~\ref{fig:bland-altman-pure_mapure} that correspond to results shown in \Cref{tab:overall_results}.

\begin{table}[H]
    \small
        \captionsetup{width=\columnwidth}
        \caption{\small{\textbf{Effect of Motion Types -- Non-rigid.} We augment UBFC-rPPG with various types of non-rigid motions (expressions) and test on the speech task, in PURE~\cite{stricker2014non}. The best results are shown in bold.}}
        \vspace{-8pt}
        \label{tab:nonrigid}
        \centering
        \small
        \setlength{\tabcolsep}{4pt}
        \adjustbox{max width=\columnwidth}{
        \begin{tabular}{llcccc}
        \toprule[1.5pt]
            & & \multicolumn{4}{c}{\textbf{Testing Set}} \\
            & & \multicolumn{4}{c}{\textbf{Non-rigid Motion Task}} \\
            \cmidrule(lr){3-6}
        \textbf{Training Set} & \textbf{Non-Rigid Motion} & MAE$\downarrow$ &  RMSE$\downarrow$ &  MAPE$\downarrow$ & $\rho$ $\uparrow$ \\ \midrule\midrule
        UBFC-rPPG & \multicolumn{1}{c}{Very Small} & 10.84 & 24.64 & 11.40 & 0.46 \\
        \grayrow
        MAUBFC-rPPG & \multicolumn{1}{c}{Small} & 1.86 & 2.79 & 2.94 & \textbf{0.99} \\
        \grayrow
        MAUBFC-rPPG & \multicolumn{1}{c}{Large} & \textbf{1.17} & \textbf{1.90} & \textbf{1.55} & \textbf{0.99} \\
        \midrule
        \multicolumn{2}{c}{\textsc{Ours vs. Baseline}} & \textbf{\textcolor{ForestGreen}{+89.21\%}} & \textbf{\textcolor{ForestGreen}{+92.29\%}} & \textbf{\textcolor{ForestGreen}{+86.40\%}} & \textbf{\textcolor{gray}{+0.00\%}} \\
        \bottomrule[1.5pt]
        \end{tabular}}
        \footnotesize
       MAE = Mean Absolute Error in HR estimation (Beats/Min), RMSE = Root Mean Square Error in HR estimation (Beats/Min), MAPE = Mean Absolute Percentage Error in HR estimation, $\rho$ = Pearson Correlation in HR estimation
\end{table}

\begin{table*}[h!]
    \small
        \captionsetup{width=\textwidth}
        \caption{\small{\textbf{Effect of Motion Types -- Rigid.} We augment UBFC-rPPG with various types of rigid head motions and test on AFRL~\cite{estepp2014recovering}. The best results are shown in bold.}}
        \vspace{-8pt}
        \label{tab:rigid}
        \centering
        \small
        \setlength{\tabcolsep}{4pt}
        \adjustbox{max width=\textwidth}{
        \begin{tabular}{llcccccccccccccccc}
        \toprule[1.5pt]
            & & \multicolumn{16}{c}{\textbf{Testing Set}} \\
            & & \multicolumn{4}{c}{\textbf{No Motion}} & \multicolumn{4}{c}{\textbf{Small Motion}} & \multicolumn{4}{c}{\textbf{Large Motion}} & \multicolumn{4}{c}{\textbf{All}} \\
            \cmidrule(lr){3-6} \cmidrule(lr){7-10} \cmidrule(lr){11-14} \cmidrule(lr){15-18}
        \textbf{Training Set} & \textbf{Rigid Motion} & MAE$\downarrow$ &  RMSE$\downarrow$ &  MAPE$\downarrow$ & $\rho$ $\uparrow$ & MAE$\downarrow$ &  RMSE$\downarrow$ &  MAPE$\downarrow$ & $\rho$ $\uparrow$ & MAE$\downarrow$ &  RMSE$\downarrow$ &  MAPE$\downarrow$ & $\rho$ $\uparrow$ & MAE$\downarrow$ &  RMSE$\downarrow$ &  MAPE$\downarrow$ & $\rho$ $\uparrow$ \\ \midrule\midrule
        UBFC-rPPG & Very Small & 1.00 & 3.86 & 1.48 & 0.95 & 2.28 & 6.36 & 3.44 & 0.85 & 7.59 & 12.91 & 10.99 & 0.49 & 4.72 & 10.01 & 6.59 & 0.67 \\
        \grayrow
        MAUBFC-rPPG & Small & \textbf{0.84} & \textbf{3.25} & \textbf{1.18} & \textbf{0.96} & \textbf{1.44} & \textbf{4.44} & \textbf{2.03} & \textbf{0.93} & 4.21 & 9.11 & 5.96 & 0.74 & \textbf{3.19} & \textbf{7.96} & \textbf{4.36} & \textbf{0.79} \\
        \grayrow
        MAUBFC-rPPG & Large & 1.00 & 3.61 & 1.37 & 0.96 & 1.78 & 5.23 & 2.49 & 0.90 & \textbf{3.64} & \textbf{8.14} & \textbf{5.12} & \textbf{0.78} & 3.39 & 8.26 & 4.58 & 0.77 \\
        \midrule
        \multicolumn{2}{c}{\textsc{Ours vs. Baseline}} & \textbf{\textcolor{ForestGreen}{+16.00\%}}  & \textbf{\textcolor{ForestGreen}{+15.80\%}} & \textbf{\textcolor{ForestGreen}{+20.27\%}} & \textbf{\textcolor{ForestGreen}{+1.05\%}} & \textbf{\textcolor{ForestGreen}{+36.84\%}} & \textbf{\textcolor{ForestGreen}{+30.19\%}} & \textbf{\textcolor{ForestGreen}{+40.99\%}} & \textbf{\textcolor{ForestGreen}{+9.41\%}} & \textbf{\textcolor{ForestGreen}{+52.04\%}} & \textbf{\textcolor{ForestGreen}{+36.95\%}} & \textbf{\textcolor{ForestGreen}{+53.41\%}} & \textbf{\textcolor{ForestGreen}{+59.18\%}} & \textbf{\textcolor{ForestGreen}{+32.42\%}} & \textbf{\textcolor{ForestGreen}{+20.48\%}} & \textbf{\textcolor{ForestGreen}{+33.84\%}} & \textbf{\textcolor{ForestGreen}{+17.91\%}} \\
        \bottomrule[1.5pt]
        \end{tabular}}
        \footnotesize
       MAE = Mean Absolute Error in HR estimation (Beats/Min), RMSE = Root Mean Square Error in HR estimation (Beats/Min), MAPE = Mean Absolute Percentage Error in HR estimation, $\rho$ = Pearson Correlation in HR estimation
\end{table*}

\begin{table*}[h!]
    \small
        \captionsetup{width=\textwidth}
        \caption{\small{\textbf{Evaluation across all datasets.} We motion-augment two training datasets, UBFC-rPPG and PURE, to create MAUBFC-rPPG and MAPURE, respectively. We observe that the motion-augmented versions produce significant improvements (shown in bold).}}
        \vspace{-8pt}
        \label{tab:overall_results}
        \centering
        \small
        \setlength{\tabcolsep}{4pt}
        \adjustbox{max width=\textwidth}{
        \begin{tabular}{llcccccccccccccccccccc}
        \toprule[1.5pt]
            & & \multicolumn{20}{c}{\textbf{Testing Set}} \\
            & & \multicolumn{4}{c}{\textbf{UBFC-rPPG}} & \multicolumn{4}{c}{\textbf{PURE}} & \multicolumn{4}{c}{\textbf{UBFC-PHYS}} & \multicolumn{4}{c}{\textbf{AFRL}} & \multicolumn{4}{c}{\textbf{MMPD}} \\
            \cmidrule(lr){3-6} \cmidrule(lr){7-10} \cmidrule(lr){11-14} \cmidrule(lr){15-18} \cmidrule(lr){19-22}
        \textbf{Training Set} & \textbf{Method} & MAE$\downarrow$ &  RMSE$\downarrow$ &  MAPE$\downarrow$ & $\rho$ $\uparrow$ & MAE$\downarrow$ &  RMSE$\downarrow$ &  MAPE$\downarrow$ & $\rho$ $\uparrow$ & MAE$\downarrow$ &  RMSE$\downarrow$ &  MAPE$\downarrow$ & $\rho$ $\uparrow$ & MAE$\downarrow$ &  RMSE$\downarrow$ &  MAPE$\downarrow$ & $\rho$ $\uparrow$ & MAE$\downarrow$ &  RMSE$\downarrow$ &  MAPE$\downarrow$ & $\rho$ $\uparrow$ \\ \midrule\midrule
        \multirow{4}{*}{Unsupervised} & Green & 19.82 & 31.49 & 18.78 & 0.37 & 10.09 & 23.85 & 10.28 & 0.34 & 13.45 & 19.11 & 16.00 & 0.31 & 7.01 & 12.52 & 9.24 & 0.52 & 16.27 & 21.74 & 20.09 & -0.04 \\
         & ICA & 14.70 & 23.71 & 14.34 & 0.53 & 4.77 & 16.70 & 4.47 & 0.72 & 8.00 & 13.51 & 9.48 & 0.48 & 6.77 & 12.25 & 8.96 & 0.51 & 13.10 & 17.84 & 16.33 & 0.03 \\
        & CHROM & 3.98 & 8.72 & 3.78 & 0.89 & 5.77 & 14.93 & 11.52 & 0.81 & 4.68 & 8.09 & 6.20 & 0.77 & 5.41 & 10.71 & 7.95 & 0.60 & 8.85 & \textbf{12.77} & 11.93 & 0.29 \\
         & POS & 4.00 & 7.58 & 3.86 & 0.92 & 3.67 & 11.82 & 7.25 & 0.88 & 4.62 & 8.02 & 6.29 & 0.78 & 6.93 & 11.89 & 10.00 & 0.49 & 8.18 & 13.04 & 11.12 & 0.31 \\ \midrule
        UBFC-rPPG & TS-CAN & - & - & - & - & 4.55 & 14.47 & 4.67 & 0.80 & 5.56 & 9.88 & 7.25 & 0.68 & 4.24 & 8.72 & 5.84 & 0.75 & 8.74 & 15.55 & 10.51 & 0.25 \\
        \grayrow
        MAUBFC-rPPG & TS-CAN & - & - & - & - & \textbf{0.96} & \textbf{4.17} & \textbf{1.13} & \textbf{0.97} & \textbf{3.93} & \textbf{7.50} & \textbf{5.24} & \textbf{0.81} & 2.67 & 6.55 & 3.65 & 0.85 & \textbf{6.80} & 14.20 & \textbf{7.97} & \textbf{0.29} \\
        PURE & TS-CAN & 1.34 & 3.01 & 1.55 & \textbf{0.99} & - & - & - & - & 4.43 & 8.12 & 5.89 & 0.78 & 2.63 & 7.35 & 3.51 & 0.82 & 8.96 & 16.59 & 10.33 & 0.15 \\
        \grayrow
        MAPURE & TS-CAN  & \textbf{1.03} & \textbf{2.70} & \textbf{1.17} & \textbf{0.99} & - & - & - & - & 4.39 & 8.10 & 5.90 & 0.78 & \textbf{2.37} & \textbf{6.28} & \textbf{3.26} & \textbf{0.87} & 8.08 & 15.38 & 9.54 & 0.18 \\
        \midrule
        \multicolumn{2}{c}{\textsc{MAUBFC-rPPG vs. UBFC-rPPG}} & - & - & - & - & \textbf{\textcolor{ForestGreen}{+78.9\%}} & \textbf{\textcolor{ForestGreen}{+71.18\%}} & \textbf{\textcolor{ForestGreen}{+75.8\%}} & \textbf{\textcolor{ForestGreen}{+21.25\%}} & \textbf{\textcolor{ForestGreen}{+29.32\%}} & \textbf{\textcolor{ForestGreen}{+24.09\%}} & \textbf{\textcolor{ForestGreen}{+27.72\%}} & \textbf{\textcolor{ForestGreen}{+19.12\%}} & \textbf{\textcolor{ForestGreen}{+37.03\%}} & \textbf{\textcolor{ForestGreen}{+24.89\%}} & \textbf{\textcolor{ForestGreen}{+37.50\%}} & \textbf{\textcolor{ForestGreen}{+13.33\%}} & \textbf{\textcolor{ForestGreen}{+22.20\%}} & \textbf{\textcolor{ForestGreen}{+8.68\%}} & \textbf{\textcolor{ForestGreen}{+24.17\%}} & \textbf{\textcolor{ForestGreen}{+16.00\%}} \\
        \midrule
        \multicolumn{2}{c}{\textsc{MAPURE vs. PURE}} & \textbf{\textcolor{ForestGreen}{+23.13\%}} & \textbf{\textcolor{ForestGreen}{+10.30\%}} & \textbf{\textcolor{ForestGreen}{+24.52\%}} & \textbf{\textcolor{gray}{+0.00\%}} & - & - & - & - & \textbf{\textcolor{ForestGreen}{+0.90\%}} & \textbf{\textcolor{ForestGreen}{+0.25\%}} & \textbf{\textcolor{darkred}{-0.17\%}} & \textbf{\textcolor{gray}{+0.00\%}} & \textbf{\textcolor{ForestGreen}{+9.89\%}} & \textbf{\textcolor{ForestGreen}{+14.56\%}} & \textbf{\textcolor{ForestGreen}{+7.12\%}} & \textbf{\textcolor{ForestGreen}{+6.10\%}} & \textbf{\textcolor{ForestGreen}{+9.82\%}} & \textbf{\textcolor{ForestGreen}{+7.29\%}} & \textbf{\textcolor{ForestGreen}{+7.65\%}} & \textbf{\textcolor{ForestGreen}{+20.00\%}} \\
        \bottomrule[1.5pt]
        \end{tabular}}
        \footnotesize
       MAE = Mean Absolute Error in HR estimation (Beats/Min), RMSE = Root Mean Square Error in HR estimation (Beats/Min), MAPE = Mean Absolute Percentage Error in HR estimation, $\rho$ = Pearson Correlation in HR estimation
\end{table*}

\begin{table*}[h!]
    \small
        \captionsetup{width=\textwidth}
        \caption{\small{\textbf{Naturalistic vs Synthetic Head Motion.} We compare the effect of adding head motions to SCAMPS and UBFC-rPPG and contrast this with using motion data in SCAMPS. Average time for augmenting each frame of a sequence is presented. The best results are shown in bold.}}
        \vspace{-8pt}
        \label{tab:synthetic_versus_real}
        \centering
        \small
        \setlength{\tabcolsep}{4pt}
        \adjustbox{max width=\textwidth}{
        \begin{tabular}{lccccccccc}
        \toprule[1.5pt]
            & \multicolumn{8}{c}{\textbf{Testing Set}} \\
            & \multicolumn{4}{c}{\textbf{PURE}} & \multicolumn{4}{c}{\textbf{AFRL}} & \multicolumn{1}{c}{Per Frame} \\
            \cmidrule(lr){2-5} \cmidrule(lr){6-9}
        \textbf{Training Set} & MAE$\downarrow$ &  RMSE$\downarrow$ &  MAPE$\downarrow$ & $\rho$ $\uparrow$ & MAE$\downarrow$ &  RMSE$\downarrow$ &  MAPE$\downarrow$ & $\rho$ $\uparrow$ & \multicolumn{1}{c}{Synthesis Time} \\ \midrule\midrule
        SCAMPS-200 (No motion) & 10.29 & 23.81 & 11.09 & 0.35 & 7.75 & 13.08 & 10.54 & 0.48 & 37s \\
        SCAMPS-200 (Motion) & 5.38 & 16.98 & 5.42 & 0.72 & 7.25 & 12.85 & 10.20 & 0.48 & 37s \\
        Wang et al.~\cite{wang2022synthetic} & 7.40 & 22.45 & 6.13 & 0.44 & - & - & - & - & N/A \\
        UBFC-rPPG & 4.55 & 14.47 & 4.67 & 0.80 & 4.72 & 10.01 & 6.59 & 0.67 & - \\
        \grayrow
        MASCAMPS-200 & 4.67 & 16.35 & 4.22 & 0.75 & 5.00 & 10.10 & 6.69 & 0.67 & \textbf{1.20s} \\
        \grayrow
        MAUBFC-rPPG & \textbf{0.96} & \textbf{4.17} & \textbf{1.13} & \textbf{0.97} & \textbf{3.24} & \textbf{7.89} & \textbf{4.37} & \textbf{0.79} & 2.39s \\
        \midrule
        \multicolumn{1}{c}{\textsc{MASCAMPS vs. SCAMPS Baseline}} & \textbf{\textcolor{ForestGreen}{+13.20\%}} & \textbf{\textcolor{ForestGreen}{+3.71\%}} & \textbf{\textcolor{ForestGreen}{+22.14\%}} & \textbf{\textcolor{ForestGreen}{+4.17\%}} & \textbf{\textcolor{ForestGreen}{+31.03\%}} & \textbf{\textcolor{ForestGreen}{+21.40\%}} & \textbf{\textcolor{ForestGreen}{+34.41\%}} & \textbf{\textcolor{ForestGreen}{+39.58\%}} & \textbf{\textcolor{ForestGreen}{+96.76\%}} \\
        \midrule
        \multicolumn{1}{c}{\textsc{MAUBFC vs. UBFC-rPPG Baseline}} & \textbf{\textcolor{ForestGreen}{+78.9\%}} & \textbf{\textcolor{ForestGreen}{+71.18\%}} & \textbf{\textcolor{ForestGreen}{+75.8\%}} & \textbf{\textcolor{ForestGreen}{+21.25\%}} & \textbf{\textcolor{ForestGreen}{+31.36\%}} & \textbf{\textcolor{ForestGreen}{+21.18\%}} & \textbf{\textcolor{ForestGreen}{+33.69\%}} & \textbf{\textcolor{ForestGreen}{+17.91\%}} & - \\
        \bottomrule[1.5pt]
        \end{tabular}}
        \footnotesize
       MAE = Mean Absolute Error in HR estimation (Beats/Min), RMSE = Root Mean Square Error in HR estimation (Beats/Min), MAPE = Mean Absolute Percentage Error in HR estimation, $\rho$ = Pearson Correlation in HR estimation, Synthesis Time = the amount of time (in seconds) it takes to synthesize a single frame, when relevant
\end{table*}

\begin{table*}[h!]
    \small
        \captionsetup{width=\textwidth}
        \caption{\small{\textbf{Effect of rPPG Estimation Models.} We train different PPG estimation networks on UBFC-rPPG and MAUBFC-rPPG and evaluate on PURE. The best results are shown in bold.}}
        \vspace{-8pt}
        \label{tab:neural_methods_ablation}
        \centering
        \small
        \setlength{\tabcolsep}{4pt}
        \adjustbox{max width=\textwidth}{
        \begin{tabular}{llcccc}
        \toprule[1.5pt]
            & & \multicolumn{4}{c}{\textbf{Testing Set}} \\
            & & \multicolumn{4}{c}{\textbf{PURE}} \\
            \cmidrule(lr){3-6}
        \textbf{Training Set} & \textbf{Method} & MAE$\downarrow$ &  RMSE$\downarrow$ &  MAPE$\downarrow$ & $\rho$ $\uparrow$ \\ \midrule\midrule
        UBFC-rPPG & DeepPhys & 5.14 & 17.20 & 4.90 & 0.72 \\
        \grayrow
        MAUBFC-rPPG & DeepPhys & 1.24 & 6.01 & 1.56 & \textbf{0.97} \\
        UBFC-rPPG & PhysNet & 8.06 & 19.71 & 13.67 & 0.61 \\
        \grayrow
        MAUBFC-rPPG & PhysNet & 2.38 & 11.29 & 2.44 & 0.88 \\
        UBFC-rPPG & TS-CAN & 4.55 & 14.47 & 4.67 & 0.80 \\
        \grayrow
        MAUBFC-rPPG & TS-CAN & \textbf{0.96} & \textbf{4.17} & \textbf{1.13} & \textbf{0.97} \\
        \bottomrule[1.5pt]
        \end{tabular}}
        \\
        \footnotesize
       MAE = Mean Absolute Error in HR estimation (Beats/Min), RMSE = Root Mean Square Error in HR estimation (Beats/Min), MAPE = Mean Absolute Percentage Error in HR estimation, $\rho$ = Pearson Correlation in HR estimation
\end{table*}

\subsection{Experimental Details}
\label{sec:details}

The predicted PPG signals were filtered using a band-pass filter with cut-offs 0.75 Hz and 2.5 Hz. The heart rate was calculated based on the predicted PPG signal using the Fast Fourier Transform (FFT), with a measurement window of the video length for UBFC-rPPG, PURE, UBFC-PHYS, and MMPD. To evaluate the AFRL dataset, a measurement window of 30 seconds was utilized for heart rate calculations. All networks were trained using an NVIDIA RTX A4500 and PyTorch~\cite{paszke2019pytorch} implementations in the publicly available rPPG-Toolbox~\cite{liu2022rppg}. All pre-processing steps and evaluation was also done in a reproducible fashion using the toolbox. The AdamW~\cite{ilya2017adamw} optimizer, a mean squared error (MSE) loss, and a cyclic learning rate scheduler was utilized with 30 epochs, a learning rate of 0.009, and a batch size of 4 for both training and inference.

For both the UBFC-rPPG dataset and PURE datasets, all subjects were augmented with motion. For our experiments, we elect to use all of the subjects in our training and train to the very last epoch. A variety of appropriately titled pre-trained models corresponding to results in the main paper and the appendices are included alongside our code in the \textit{pretrained\_models} folder.

For \Cref{tab:synthetic_versus_real}, we consider 200 samples from the SCAMPS dataset that consist of significant synthetically generated rigid and non-rigid head motion (ID 1801 to 2000) as SCAMPS-200 (Motion). We then take instances from the SCAMPS dataset with no head motion (ID 1 to 200) and augment them with naturalistic head motion using our motion synthesis pipeline and a subset of driving videos from the TalkingHead-1KH dataset to produce MASCAMPS-200. We choose driving videos with a range of mean standard deviation in AUs from 0.35 to 0.40 intensity and a range of mean standard deviation in head pose rotations from 0.05 to 0.125 rad. Note that both SCAMPS-200 (Motion) and MASCAMPS-200 consist of synthetics with the same number of identities, with the only difference being synthetic and naturalistic head motion, respectively.

\subsection{Effect of Multiple Augmentations}
\label{sec:m_scale}

We consider whether it is plausible to augment the same source video with multiple driving videos using neural motion transfer. Thus, the newly augmented dataset has the same number of identities as the original dataset but a significantly larger variation in motions. Our goal is to analyze how many times one can augment a single source video before the performance starts to saturate or drop. 

We consider UBFC-rPPG as training data that we augment with randomly sampled driving videos from the TalkingHead-1K dataset to produce MAUBFC-rPPG. We augment the same source video from 1 to 4 times with different driving videos and evaluate on the PURE~\cite{stricker2014non} dataset and the UBFC-PHYS~\cite{meziatisabour2021ubfc} dataset and report the results in \Cref{tab:main_maubfc_scaling}. We notice that the results saturate pretty quickly and can start to decline after augmenting more than 2 times. This is presumably due to the fact that we were not augmenting other aspects of the subjects' appearance (e.g., skin tone, identity, etc.).

\section{Intra-dataset  Results}
\label{sec:intradataset}

We include intra-dataset results not included in the main paper here for reference. We utilize all of the tasks from the PURE dataset. We train on subjects 1, 2, 3, 4, and 5 and then test on subjects 6, 7, 8, 9, and 10. We then train on subjects 6, 7, 8, 9, and 10 and then test on subjects 1, 2, 3, 4, and 5. We average the results from these two experiments and repeat the aforementioned process for the motion-augmented version of PURE. We find that motion augmentation helps as an intra-dataset augmentation technique.

\begin{table}[H]
    \small
        \captionsetup{width=\columnwidth}
        \caption{\textbf{PURE Intra-dataset Results.} We use motion augmentation to augment half of the subjects in the PURE dataset at a time, while testing on the corresponding other half. The averaged results are shown below, with the best result in each column bolded.}
        \vspace{-8pt}
        \label{tab:pure_intradataset}
        \centering
        \small
        \setlength{\tabcolsep}{4pt}
        \adjustbox{max width=\columnwidth}{
        \begin{tabular}{lcccc}
        \toprule[1.5pt]
            & \multicolumn{4}{c}{\textbf{Testing Set}} \\
            & \multicolumn{4}{c}{\textbf{PURE}} \\
            \cmidrule(lr){2-5}
        \textbf{Training Set} & MAE$\downarrow$ &  RMSE$\downarrow$ &  MAPE$\downarrow$ & $\rho$ $\uparrow$ \\ \midrule\midrule
        PURE & 2.52 & 8.92 & 2.55 & 0.92 \\
        \grayrow
        MAPURE & \textbf{1.61} & \textbf{5.50} & \textbf{1.77} & \textbf{0.97} \\
        \midrule
        \multicolumn{1}{c}{\textsc{Ours vs. Baseline}} & \textbf{\textcolor{ForestGreen}{+36.1\%}} & \textbf{\textcolor{ForestGreen}{+38.34\%}} & \textbf{\textcolor{ForestGreen}{+30.59\%}} & \textbf{\textcolor{ForestGreen}{+5.43\%}} \\
        \bottomrule[1.5pt]
        \end{tabular}}
        \footnotesize
       MAE = Mean Absolute Error in HR estimation (Beats/Min), RMSE = Root Mean Square Error in HR estimation (Beats/Min), MAPE = Mean Absolute Percentage Error in HR estimation, $\rho$ = Pearson Correlation in HR estimation
\end{table}

\begin{table}[h!]
    \small
        \captionsetup{width=\columnwidth}
        \caption{\small{\textbf{Effect of Multiple Augmentations.} Augmenting each source video of UBFC-rPPG 1x, 2x, 3x, and 4x, we test on PURE and UBFC-PHYS datasets. The best results are shown in bold.}}
        \vspace{-8pt}
        \label{tab:main_maubfc_scaling}
        \centering
        \small
        \setlength{\tabcolsep}{4pt}
        \adjustbox{max width=\columnwidth}{
        \begin{tabular}{lllcccc}
        \toprule[1.5pt]
            & & & \multicolumn{4}{c}{\textbf{Testing Set}} \\
            & & & \multicolumn{2}{c}{\textbf{PURE}} & \multicolumn{2}{c}{\textbf{UBFC-PHYS}} \\
            \cmidrule(lr){4-5} \cmidrule(lr){6-7}
        \textbf{Training Set} & \textbf{Size} & \textbf{Subjects} & MAE$\downarrow$ & MAPE$\downarrow$ & MAE$\downarrow$ &  MAPE$\downarrow$ \\ \midrule\midrule
        UBFC-rPPG & \multicolumn{1}{c}{42} & \multicolumn{1}{c}{42} & 4.55 & 4.67 & 5.56 & 7.25 \\
        MAUBFC-rPPG & \multicolumn{1}{c}{42} & \multicolumn{1}{c}{42} & 0.96 & 1.13 & 3.93 & 5.24 \\
        \grayrow
        MAUBFC-rPPG 2x & \multicolumn{1}{c}{84} & \multicolumn{1}{c}{42} & 0.94 & 1.10 & \textbf{3.90} & \textbf{5.22} \\
        \grayrow
        MAUBFC-rPPG 3x  & \multicolumn{1}{c}{126} & \multicolumn{1}{c}{42} & \textbf{0.92} & \textbf{1.09} & 3.97 & 5.31 \\
        \grayrow
        MAUBFC-rPPG 4x & \multicolumn{1}{c}{168} & \multicolumn{1}{c}{42} & 1.02 & 1.25 & 4.10 & 5.40 \\
        \midrule
        \multicolumn{3}{c}{\textsc{Ours vs. Baseline}} & \textbf{\textcolor{ForestGreen}{+2.63\%}} & \textbf{\textcolor{ForestGreen}{+2.31\%}} &  \textbf{\textcolor{ForestGreen}{+0.76\%}} &  \textbf{\textcolor{ForestGreen}{+0.38\%}} \\
        \bottomrule[1.5pt]
        \end{tabular}}
    \vspace{-0.2cm}
\end{table}

\section{Motion Augmented rPPG Videos}
\label{sec:motion_augmented_videos}

We provide code for augmenting various camera physiology datasets and pre-trained models trained on motion-augmented data. Additionally, we provide various files to easily train on baselines and motion augmented data using the publicly available rPPG-Toolbox~\cite{liu2022rppg}. Pre-trained models using the baseline UBFC-rPPG or PURE datasets can be found in the rPPG-Toolbox. We also include motion analysis scripts that utilize OpenFace~\cite{baltruvsaitis2016openface} to analyze both rigid and non-rigid motion in videos and generate plots. All of these additional materials be found through our project page: \url{https://motion-matters.github.io/}.

\subsection{The Effect of Motion Transfer on PPG}
\label{sec:ma_effects}

\begin{figure*}[h!]
    \begin{center}
    \includegraphics[width=\textwidth]{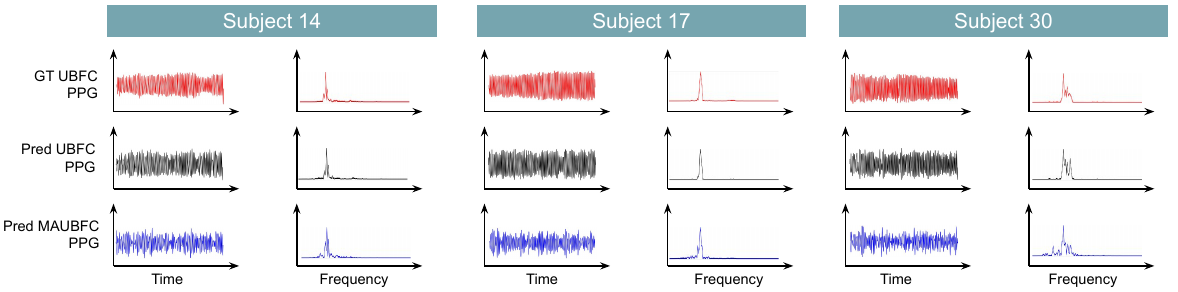}
    \end{center}
    \vspace{-0.5cm}
    \caption{\small{\textbf{Signal Prediction on UBFC-rPPG and MAUBFC-rPPG.} We provide three, subject-wise examples of signal prediction using a neural method, TS-CAN, on unaugmented and motion-augmented videos from the UBFC-rPPG dataset.}}
    \label{fig:example_signals}
    \vspace{-0.25cm}
    \end{figure*}

In \Cref{fig:example_signals}, we provide additional qualitative examples of the effect of motion transfer on the underlying PPG signal in rPPG videos. All examples include a plot of the gold-standard PPG signal, the predicted PPG signal from the unaugmented source rPPG video by a TS-CAN model, and the predicted PPG signal from the motion-augmented source rPPG video by a TS-CAN model. The TS-CAN models utilized are trained on a larger superset of the shown examples (e.g., UBFC, MAUBFC) with the same experimental settings mentioned in Section~\ref{sec:details}. As shown by these qualitative examples, a neural method such as TS-CAN is capable of recovering the underlying PPG signal despite the application of motion augmentation.

Our observation that the underlying physiological signal is reasonably preserved after motion augmentation and, therefore, useful as training data may seem contradictory to prior works (e.g., ~\cite{ciftci2020fakecatcher}) that analyze the loss of physiological information as a means to identify deep fake videos. The methods utilized in such prior works typically animate a single frame or swap faces, which we would respectively expect to not have and not preserve a useful rPPG signal. When a method that does animate a whole video without destroying the video subject identity is mentioned, the source videos in question can be highly compressed videos rather than raw, uncompressed videos from rPPG datasets that we used and are expected to have a useful rPPG signal to begin with. It is well-known that various degrees of compression in videos can severely degrade the underlying rPPG signal~\cite{mcduff2017impact}.

\section{Datasets} 
\label{sec:datasets}

\begin{table}[h!]
    \small
        \captionsetup{width=\columnwidth}
        \caption{A Summary of the rPPG Benchmark Datasets.}
        \vspace{-8pt}
        \label{tab:data_desp}
        \centering
        \small
        \setlength{\tabcolsep}{4pt}
        \adjustbox{max width=\columnwidth}{
        \begin{tabular}{lcc}
        \multicolumn{3}{c}{\includegraphics[width=0.7\textwidth]{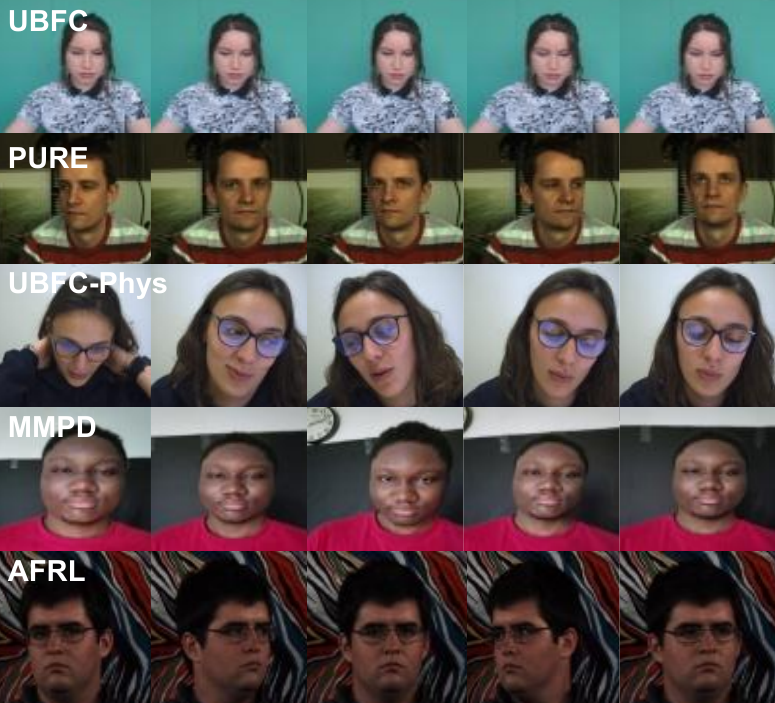}} \\
        \toprule[1.5pt]
        \textbf{Dataset} & \textbf{Subjects / Videos} & \textbf{Motion Tasks}   \\ \midrule\midrule
        UBFC-rPPG & 42 / 42 & Stationary \\
        PURE & 10 / 59 & Stationary, Talking, Rotation, Translation \\
        UBFC-Phys & 56 / 168 & Stationary, Talking, Head Rotation \\
        MMPD & 33 / 660 & Stationary, Talking, Walking, Head Rotation \\
        AFRL & 25 / 300 & Stationary, Head Rotation \\
        \midrule
        \bottomrule[1.5pt]
        \end{tabular}}
    \vspace{-0.6cm}
\end{table}

\subsection{Source Videos for Motion Synthesis}
\label{sec:source_datasets}

We use the following state-of-the-art datasets for source videos used in our motion synthesis pipeline:

\textbf{UBFC-rPPG}~\cite{bobbia2019unsupervised}: The UBFC-rPPG video dataset contains 42 RGB videos, one per subject, at 30 Hz. The videos were collected with a Logitech C920 HD Pro with a resolution of 640x480 and a CMS50E transmissive pulse oximeter was utilized in order to record gold-standard PPG signals. The UBFC-rPPG dataset contains minimal motion, with subjects being asked to simply sit one meter away from the camera in an environment with both artificial and natural lighting. When utilized as source videos, we utilized all videos from the UBFC-rPPG dataset. When utilized for evaluation, we also utilized all videos from the UBFC-rPPG dataset.

\textbf{PURE}~\cite{stricker2014non}: The PURE dataset contains 59 videos, each corresponding to a unique task, per a subject. The six tasks involve staying steady, talking, slow head translation, fast head translation, small head rotation, and medium head rotation. There are 10 subjects total with subject 6's talking task video being excluded. All of the videos were captured with an RGB eco274CVGE camera (SVS-Vistek GmbH) at a resolution of 640x480 and 60 Hz.  During all tasks, the subject was asked to be seated in front of the camera at an average distance of 1.1 meters and lit from the front with ambient natural light through a window. A gold-standard measure of PPG was collected with a pulse oximeter, CMS50E, attached to the finger. When utilized as source videos, we utilized all videos from the PURE dataset. When utilized for evaluation, we also utilized all videos from the PURE dataset.

\textbf{SCAMPS}~\cite{mcduff2022scamps}: The SCAMPS dataset contains 2,800 synthetic videos that were generated using a blendshape-based rig with 7,667 vertices and 7,414 polygons, with distinct identities being learned from a set of high-quality facial scans. Blood flow, and subsequently the underlying physiological signals, are simulated using the modification of physically-based shading materials. The SCAMPS dataset contains a variety of rigid and non-rigid head motions, with varying intensities. The dataset also contains a variety of lighting conditions and background conditions. Each SCAMPS video is 20 seconds in length, with 600 frames at a sampling rate of 30 Hz. We only utilize portions of the SCAMPS dataset as source videos in our ablation study regarding synthetic versus naturalistic head motion.

\subsection{Driving Videos for Motion Synthesis}
\label{sec:driving_datasets}

We use the following datasets for driving videos used in our motion synthesis pipeline:

\textbf{TalkingHead-1KH}~\cite{Wang2020OneShotFN}: The TalkingHead-1KH dataset is a publicly available, large-scale talking-head video dataset used as a benchmark for Face-Vid2Vid~\cite{Wang2020OneShotFN} and entirely sourced from YouTube videos. It contains 180K unconstrained videos of people speaking in a variety of real-world contexts, leading to a rich diversity in both rigid and non-rigid motion. The videos are of varied resolutions, but there is an emphasis on collecting high quality, high resolution videos which compose a significant portion of the dataset (with a resolution of at least 512x512). We elect to filter the dataset for head pose such that any videos where the head pose, on average, is outside +/- 20 degrees are removed. This prevents damaging motion augmentation artifacts due to impractical differences in the head pose in the source video and the head pose in the driving videos, but comes at the cost of reduced head pose variations. We also filter by facial action units (AUs) (0 to 5, in units of intensity) such that any videos below a mean standard deviation in facial
AUs of 0.15 is filtered out. This prevents driving videos that are not suitable for our application from being used - for example, a driving video that is effectively a slide show and doesn't have any naturalistic motion upon qualitative inspection.

\textbf{CDVS}: The CDVS contains 90 self-captured videos by 5 subjects with heavily constrained, unnatural motion used only for ablation studies to understand the impact of augmenting data with various degrees of rigid and non-rigid motion. Subjects self-capture the videos in a variety of settings with artificial lighting of the face in an indoors setting. When capturing a video to show one of the two types of motion we study, subjects are asked to constrain the other motion type as much as possible. The CDVS will be released in the future for research purposes.

\subsection{Additional Datasets for Evaluation}
\label{sec:eval_datasets}

In addition to using UBFC-rPPG~\cite{bobbia2019unsupervised} and PURE~\cite{stricker2014non} as both source video datasets in the motion synthesis pipeline and datasets for evaluation, we use three additional state-of-the-art datasets for evaluation:

\textbf{UBFC-PHYS}~\cite{meziatisabour2021ubfc}: The UBFC-PHYS dataset is a multimodal dataset with 168 RGB videos, with 56 subjects (46 women and 10 men) per a task. There are three tasks with significant amounts of both rigid and non-rigid motion - a rest task, a speech task, and an arithmetic task. Gold-standard BVP and electrodermal activity (EDA) measurements were collected via the Empatica E4 wristband. The videos were recorded at a resolution of 1024x1024 and 35Hz with a EO-23121C RGB digital camera. We utilized all of the tasks and the same subject sub-selection list provided by the authors of the dataset in the second supplementary material of Sabour et al.~\cite{meziatisabour2021ubfc} for evaluation. This means we eliminated 14 subjects (s3, s8, s9, s26, s28, s30, s31, s32, s33, s40, s52, s53, s54, s56) for the rest task, 30 subjects (s1, s4, s6, s8, s9, s11, s12, s13, s14, s19, s21, s22, s25, s26, s27, s28, s31, s32, s33, s35, s38, s39, s41, s42, s45, s47, s48, s52, s53, s55) for the speech task, and 23 subjects (s5, s8, s9, s10, s13, s14, s17, s22, s25, s26, s28, s30, s32, s33, s35, s37, s40, s47, s48, s49, s50, s52, s53) for the arithmetic task.

\textbf{AFRL}~\cite{estepp2014recovering}: The AFRL dataset contains 300 videos of 25 participants (17 males, 8 females) recorded at 658x492 resolution and 120 FPS. Gold-standard physiological signals were measured using the fingertip reflectance PPG method. Participants were asked to perform a series of tasks, resulting in 12 tasks total. With a black background behind the participant, the tasks entailed sitting still with a chin-rest, sitting still without a chin rest, rotating the head with an angular velocity of 10 degrees/second, 20 degrees/second, and 30 degrees/second, and finally randomly orienting their head once per a second to a predefined location. This resulted in six recordings, which were repeated once with a colorful background, resulting in 12 videos per a participant. As a part of our pre-processing steps for AFRL, we down-sampled the videos to 30 FPS. We utilized all of the videos for evaluation.

\textbf{MMPD}~\cite{tang2023mmpd}: The Multi-domain Mobile Video Physiology Dataset (MMPD) dataset contains 11 hours of recordings from mobile phones of 33 subjects. Gold-standard PPG signals were simultaneously recorded using an HKG-07C+ oximeter. The dataset was designed to capture variations in skin tone, body motion, and lighting conditions in videos useful for the rPPG task. Videos were collected under three artificial light sources: i) low LED light (100 lumens on the face region), ii) mid-level incandescent light (200 lumens on the face region), and iii) high LED light (300 lumens on the face region). Videos were also collected under natural light, which varied from 300-800 lumens intensity on the face region. Videos were recorded following an experimental procedure in which participants performed a variety of tasks in different lighting conditions - a stationary task, a head rotation task, a talking task, and a walking task. We evaluated on videos with artificial lighting, Fitzpatrick scale skin tone type 3, and any of the four tasks (stationary, head rotation, talking, and walking) that correspond to varying degrees of rigid and non-rigid motion.

\section{Broader Impact Statement}
\label{sec:broaderimpact}

While generating synthetic videos that are indistinguishable from those of real people has concerning use cases, there are positive applications of this technology can enabled.  In the medical domain simulators are increasingly being tested within specific applications~\cite{hernandez2022synthetic,hahn2022contribution}. It is important that the limitations of generative models are understood as these may impact the performance of the resulting models trained using simulated data. It is possible for generative approaches to compound harmful biases~\cite{maluleke2022ganbias} and motion augmentation algorithms can be used for troubling negative applications. To mitigate negative outcomes, we license our source code using responsible behavioral use licenses used across a large number of publicly released machine learned models~\cite{contractor2022behavioral}.

\clearpage

{\small
\bibliographystyle{ieee_fullname}
\bibliography{references}
}

\end{document}